\documentclass[10pt,journal]{IEEEtran}

\usepackage{amsmath,amsfonts,amssymb,bm} 
\usepackage{pifont}          
\usepackage{tikz,graphicx,overpic}      
\usepackage{array,booktabs,multirow,makecell} 
\usepackage[table]{xcolor}              
\usepackage{arydshln}                   
\usepackage{ragged2e}        
\usepackage{algorithmic}    
\usepackage{enumerate}                  
\usepackage[compress]{cite}           
\usepackage[colorlinks,linkcolor=red,anchorcolor=blue,citecolor=green]{hyperref} 
\usepackage{cuted}
\usepackage{capt-of}

\newcommand{\cmark}{\textcolor{green!70!black}{\ding{51}}}
\newcommand{\xmark}{\textcolor{red}{\ding{55}}}

\definecolor{darkgreen}{rgb}{0.0, 0.5, 0.0}
\definecolor{lightblue}{rgb}{0.93,0.95,1.0}
\definecolor{lightshade}{rgb}{0.9,0.9,0.9}

\definecolor{globalbg}{RGB}{238,246,232}
\definecolor{globaltext}{RGB}{72,125,48}

\definecolor{localbg}{RGB}{255,243,228}
\definecolor{localtext}{RGB}{202,91,8}

\definecolor{dualbg}{RGB}{252,232,228}
\definecolor{dualtext}{RGB}{190,72,61}

\definecolor{totalbg}{RGB}{242,242,242}
\definecolor{headerbg}{RGB}{242,242,242}
\definecolor{sfxlbg}{RGB}{242,246,250}
\definecolor{globalroute}{HTML}{4D7F2F}
\definecolor{localroute}{HTML}{D56300}
\definecolor{dualroute}{HTML}{C3473D}

\definecolor{third}{rgb}{1, 1, 0.7}
\definecolor{second}{rgb}{1, 0.85, 0.7}
\definecolor{first}{rgb}{1, 0.7, 0.7}
\newcommand{\fs}{\cellcolor{first}}   
\newcommand{\nd}{\cellcolor{second}}      
\newcommand{\rd}{\cellcolor{third}}      

\graphicspath{{./figs/}}
\DeclareGraphicsExtensions{.pdf,.jpeg,.png,.eps}

\newcommand{\resubmit}[1]{{\textcolor{black}{#1}}}

\begin{document}
\title{\resubmit{AdaptVPR: Route-Aware Hard Positive Generation for \\ Robust Visual Place Recognition}}

\author{
Shunpeng~Chen$^{\ast}$,~Jingyi~Zhang$^{\ast}$,~Changwei~Wang,~Shengpeng~Xu,~Yukun~Song,
Xingtian~Pei,~Jinzhou~Lin,~Li~Guo,~and~Shibiao~Xu$^{\dag}$
\thanks{Shunpeng Chen, Jingyi Zhang, Shengpeng Xu, Yukun Song, Xingtian Pei, Jinzhou Lin, Li Guo, and Shibiao Xu are with the School of Artificial Intelligence, Beijing University of Posts and Telecommunications.}
\thanks{Changwei Wang is with the Key Laboratory of Computing Power Network and Information Security, Ministry of Education, Shandong Computer Science Center, Qilu University of Technology.}
\thanks{$^{\ast}$Shunpeng Chen and Jingyi Zhang contributed equally to this work (e-mail: \{shunpengchen, jingyizhang18\}@bupt.edu.cn).}
\thanks{$^{\dag}$Corresponding author: Shibiao Xu (e-mail: shibiaoxu@bupt.edu.cn).}
}

\markboth{Journal of \LaTeX\ Class Files,~Vol.~14, No.~8, August~2021}%
{Shell \MakeLowercase{\textit{et al.}}: A Sample Article Using IEEEtran.cls for IEEE Journals}


\maketitle

\begin{abstract}
\justifying
Visual Place Recognition (VPR) localizes a query image by retrieving database images of the same or nearby place, yet its robustness is often degraded by domain shifts arising from illumination, weather, seasonal changes, and dynamic occlusions. One contributing factor is the limited appearance diversity of the same place in existing training data.
To address this issue, we propose AdaptVPR, a route-aware generative augmentation framework that constructs same-place hard positives for robust VPR training. AdaptVPR first uses a vision language model to parse scene attributes and estimate editing feasibility, while a rule-based scheduler determines the generation route according to editability scores and risk constraints. The generation process is decomposed into three complementary routes: the Global Appearance Route introduces global scene changes in weather, illumination, and time of day; the Local Occlusion Route inserts plausible dynamic occluders; and the Dual Route combines both types of perturbations to produce more challenging appearance shifts. Each generated candidate is evaluated using a VPR-oriented verification scheme based on geometric consistency and appearance diversity, reducing the risk of structural drift while ensuring sufficient appearance variation. Global candidates are generated once and rejected if verification fails, while Local Occlusion and Dual candidates use verification feedback for limited prompt refinement and regeneration. Using this framework, we construct AdaptCities, containing 160K verified synthetic same-place hard positives.
Experiments across multiple VPR baselines and vision foundation backbones show consistent gains on standard benchmarks and substantial improvements under challenging domain shifts, with R@1 gains of up to 9.2\%.
The source code and data resources are publicly available at \url{https://github.com/chenshunpeng/AdaptVPR}.

\end{abstract}

\begin{IEEEkeywords}
Visual Place Recognition, Hard Positive Generation, Diffusion Models, Domain Shift, Data Augmentation
\end{IEEEkeywords}

\section{Introduction}
\label{sec:introduction}

Visual Place Recognition (VPR) is a fundamental component of long term visual localization~\cite{fan2022learning,zou2012coslam,brachmann2021visual}, autonomous driving~\cite{chitta2022transfuser,teng2026deep,zhang2026benchmarking}, and mobile robot navigation~\cite{liu2024integrating,xu2024local}. Given a query image, a VPR system retrieves database images captured at the same or nearby locations, thereby formulating localization as a large scale image retrieval problem. Over the past decade, VPR has evolved from handcrafted local features and aggregation based representations~\cite{SIFT,SURF,BoW,vlad2010} to deep global descriptors~\cite{netvlad,delg,patchvlad,ali2023global,leyva2023data,structvpr++}, classification based geo-localization~\cite{cosplace,eigenplaces,Divide_classify,MutualVPR}, and more recently, representation learning built upon vision foundation models~\cite{anyloc,superplace,megaloc,Unipr-3d,A2GC}. Despite these advances, robust place recognition under open world domain shifts remains challenging. Images of the same place may undergo changes caused by illumination, weather, season, traffic, and dynamic occlusions. Such variations can alter visual appearance more strongly than the underlying place identity itself, making it difficult for retrieval models to learn representations that remain stable beyond the training distribution.

\begin{figure*}[t]
\centering
\includegraphics[width=\textwidth]{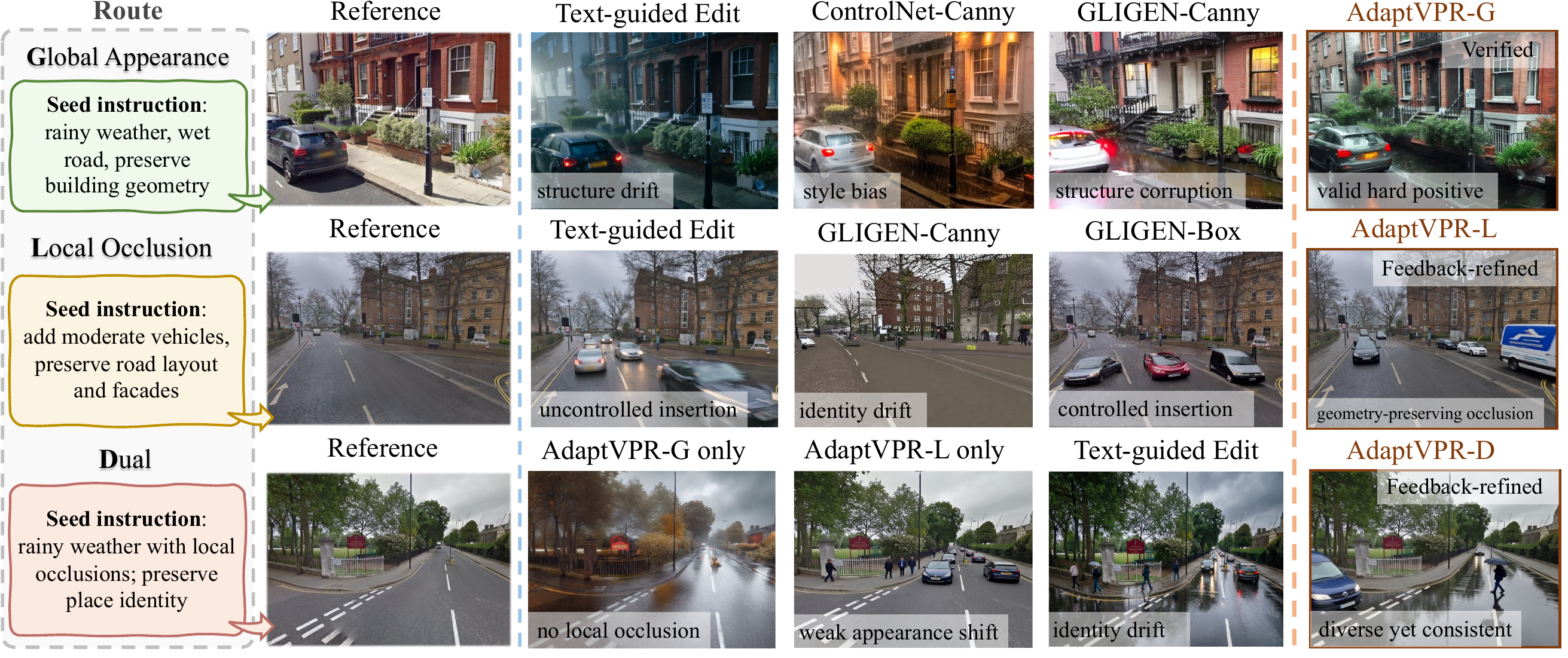}
\caption{
VPR-oriented hard positive generation.
Given the same reference image and seed instruction, generic editing methods may introduce structure drift, identity drift, or uncontrolled object insertion.
AdaptVPR decomposes generation into the Global Appearance Route, Local Occlusion Route, and Dual Route, and uses VPR-oriented verification to retain visually diverse yet geometrically consistent same-place positives.
For the Local Occlusion Route and Dual Route, verification feedback further guides prompt refinement through a reflective generation loop.
}
\label{fig:intro_vis}
\vspace{-0.3cm}
\end{figure*}

A natural way to improve such robustness is to expose VPR models to more diverse observations of the same place during training. However, collecting real revisits that cover combinations of weather, illumination, seasonal conditions, and dynamic foreground changes is expensive and difficult to scale. Conventional image augmentations, such as color jittering, cropping, blurring, and random erasing, provide useful low level perturbations but cannot faithfully reproduce complex domain shifts such as rainy nights, snowy roads, strong reflections, dynamic traffic, or structured foreground occlusions. Generative augmentation therefore provides an appealing alternative. Early image translation approaches, including CycleGAN and ToDayGAN~\cite{CycleGAN,ToDayGAN}, have been used to transfer appearance across adverse conditions, while recent VPR studies further explore nighttime translation and synthetic street view generation~\cite{Npr,DiffPlace}. The emergence of diffusion models further expands this space by enabling flexible text and condition controlled editing~\cite{DDPM,LDM,ControlNet,Instructpix2pix}.

\begin{figure}[t]
    \centering
    \includegraphics[width=\linewidth]{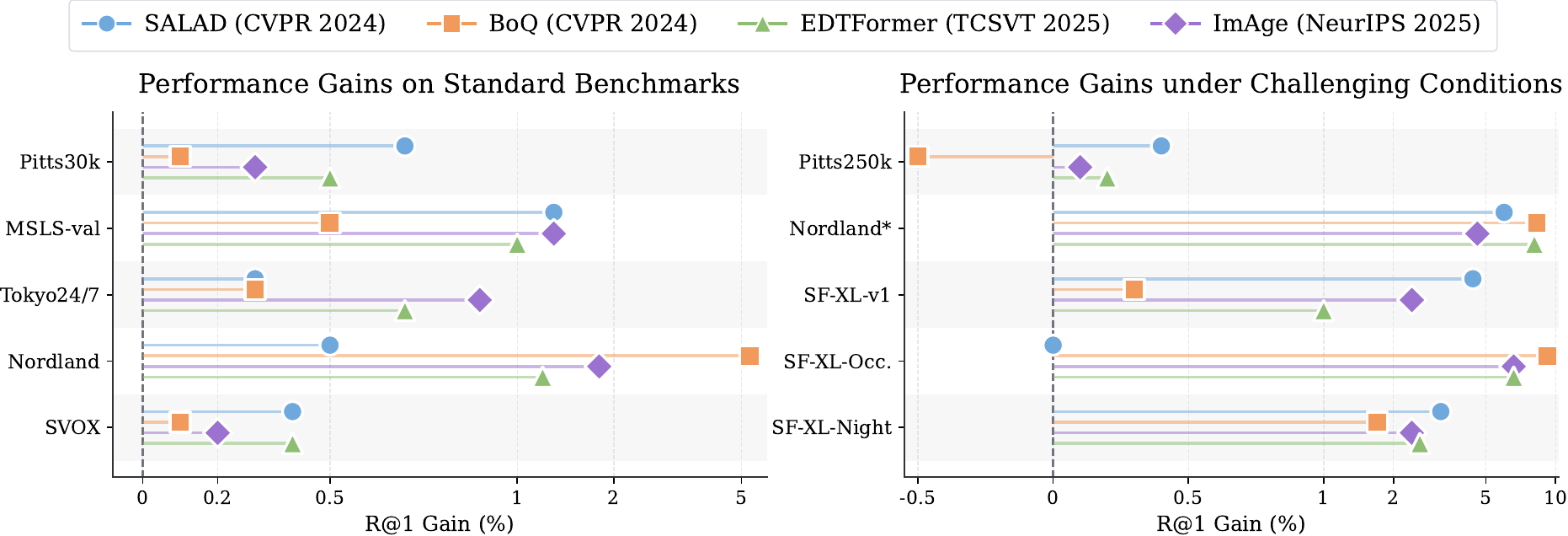}
    \caption{R@1 gains from AdaptVPR across representative VPR methods.}
    \label{fig:r1_gains}
    \vspace{-0.2cm}
\end{figure}

However, generative augmentation for VPR faces a requirement that differs fundamentally from generic image synthesis. A useful generated image should differ sufficiently from its source to provide an informative hard positive, while preserving the spatial structures that define the physical place. Optimizing only for realism, text alignment, or editability does not guarantee this property. A visually plausible rainy street or a realistic vehicle insertion may still modify building facades, road topology, lane markings, window layouts, or scene boundaries, thereby changing cues that are critical for place recognition~\cite{DiffPlace,sferrazza2025match}. Similarly, diffusion models conditioned on layouts, boxes, edges, or depth provide stronger spatial control but are not explicitly designed to preserve place identity~\cite{ControlNet,GLIGEN,LayoutEdit,FreeFine}. 
If structurally corrupted images are treated as positives during metric learning, they may act as false positives and introduce undesirable supervision into the learned embedding space. As illustrated in Fig.~\ref{fig:intro_vis}, generic editing strategies may therefore produce realistic images that are nevertheless unsuitable for VPR training. The challenge is consequently not to generate more images, but to generate valid same-place hard positives that introduce meaningful appearance changes while limiting place identity drift.

Many existing generative augmentation pipelines largely follow a feed-forward process in which candidates are generated and subsequently selected or filtered~\cite{Npr,QdaVPR,UGNAVPR,DiffPlace,gift}. This passive generation paradigm leaves two important issues insufficiently addressed. First, different forms of domain shift require different editing behaviors. Weather, illumination, and time of day should affect the scene globally while preserving its spatial layout, whereas vehicles, pedestrians, and other dynamic occluders should modify localized regions without rewriting the background geometry. Handling these factors through a single static generation strategy can lead to insufficient appearance change, excessive structural modification, or implausible object insertion. Second, existing pipelines typically do not exploit failure-specific verification feedback to iteratively correct unsuccessful candidates. For localized or compound edits, however, verification signals can indicate whether the current generation is too conservative, too aggressive, or geometrically inconsistent. This observation motivates a generation process that combines scene dependent editing strategies with VPR specific feedback before a synthetic image is admitted into the training set.

Based on these observations, we argue that robustness under domain shift can be improved by explicitly expanding the same-place appearance distribution. The same geographical place should be observed under substantially different visual conditions while remaining close in the learned descriptor space. Following this perspective, we propose \textbf{AdaptVPR}, a route-aware generative augmentation framework for constructing same-place hard positives. AdaptVPR first uses a vision language model to parse scene attributes and estimate the feasibility of different edits. A rule-based scheduler then determines the executable generation route according to editability scores and risk constraints. The generation process is decomposed into three complementary routes. 
The \textcolor{globalroute}{\textbf{Global Appearance Route}} introduces global variations in weather, illumination, and time of day. The \textcolor{localroute}{\textbf{Local Occlusion Route}} inserts plausible dynamic occluders while preserving the global scene. The \textcolor{dualroute}{\textbf{Dual Route}} combines both types of perturbations to construct more challenging compound shifts.
This decoupling allows the generation strategy to better match the physical characteristics of different visual changes instead of forcing all samples through a single editing process.

To reduce the risk of introducing harmful positives, each generated candidate is evaluated using a geometry and diversity verification scheme designed for VPR. Geometric consistency is estimated from local feature correspondences and robust geometric fitting, providing a proxy signal for structural preservation. Appearance diversity evaluates whether the intended visual change is sufficiently informative. The verification policy is adapted to the characteristics of each generation route. AdaptVPR further employs selective reflection. The Global Appearance Route performs a single generation followed by verification and direct rejection if the candidate is invalid. In contrast, candidates from the Local Occlusion Route and Dual Route can use verification feedback to refine their prompts and regenerate within a limited reflection budget.

Using AdaptVPR, we construct \textbf{AdaptCities} from GSV-Cities~\cite{gsv}\footnote{GSV-Cities~\cite{gsv} is a curated VPR training dataset comprising diverse Google Street View images collected across 23 cities and is available at \href{https://github.com/amaralibey/gsv-cities}{https://github.com/amaralibey/gsv-cities}.}, containing 160K verified synthetic same-place hard positives covering global appearance changes, local occlusions, and their combinations.
AdaptVPR operates entirely at the training data level and can be integrated into different VPR models as a general augmentation strategy. 
As shown in Fig.~\ref{fig:r1_gains}, AdaptVPR consistently improves multiple representative VPR methods across ten benchmark datasets, with particularly pronounced gains under challenging domain shifts.

Our contributions are summarized as follows:
\begin{itemize}

    \item We formulate generative augmentation for VPR as the construction of same-place hard positives, emphasizing the need to expand the diversity of same-place appearances while controlling identity drift under domain shift.

    \item We propose AdaptVPR, a route-aware generative augmentation framework that combines VLM based scene understanding, rule-based route scheduling, and three complementary generation routes for global appearance changes, local occlusions, and compound domain shifts.

    \item We introduce a route-specific geometry and diversity verification scheme together with selective reflection, where verification feedback is used to refine Local and Dual generations while invalid Global candidates are rejected.

    \item We construct AdaptCities with 160K verified synthetic same-place hard positives and demonstrate the generality of AdaptVPR across VPR baselines and vision foundation backbones, yielding consistent retrieval gains and strong robustness improvements under domain shifts.

\end{itemize}

\section{Related Work}
\label{sec:related work}
\subsection{Visual Place Recognition}

Early Visual Place Recognition (VPR) methods relied primarily on handcrafted local descriptors and aggregated them into image-level representations using Bag-of-Words, VLAD, or Fisher Vector~\cite{arandjelovic2013all,vlad2010,BoW,Fisher_Vector,densevlad}. With the development of deep learning, CNN-based feature extraction and learnable aggregation became the dominant paradigm, represented by NetVLAD and its variants~\cite{netvlad,Nextvlad,speNetvlad,patchvlad}, followed by architectures such as MixVPR~\cite{mixvpr} and BoQ~\cite{boq} that improve feature interaction and global descriptor aggregation. Another line of work introduces classification-based training strategies to improve scalability and representation learning. CosPlace~\cite{cosplace} formulates VPR training as classification over geographical groups, while EigenPlaces~\cite{eigenplaces} constructs viewpoint-aware training classes to learn more robust global descriptors. Divide\&Classify~\cite{Divide_classify} further investigates classification-based inference for city-wide localization.

More recently, vision foundation models, particularly DINOv2~\cite{dinov2}, have substantially advanced VPR representations. AnyLoc~\cite{anyloc} demonstrates strong zero-shot place recognition by exploiting pretrained foundation features, while subsequent methods adapt or aggregate these representations for VPR-specific objectives. SALAD~\cite{salad} reformulates local feature aggregation through optimal transport, and CliqueMining~\cite{izquierdo2024close} improves geographic distance sensitivity by mining visually related image cliques during training. SuperVLAD~\cite{supervlad} simplifies VLAD aggregation with substantially fewer clusters and improves cross-domain generalization. SelaVPR and SelaVPR++~\cite{selavpr,selavpr++} develop parameter-efficient adaptation and retrieval strategies for foundation models, while FoL and FoL++~\cite{FoL,FoL++} exploit discriminative spatial regions and adaptive re-ranking to improve robustness and efficiency. Recent approaches further explore stronger training and aggregation strategies; for example, ImAge~\cite{image} incorporates learnable tokens into Transformer representations, whereas SAGE~\cite{sage} combines local feature adaptation with an online geo-visual graph and adaptive hard sample mining.
DialogueVPR \cite{DialogueVPR} extends VPR toward conversational localization through iterative dialogue-based reasoning.
EfficientVPR \cite{EfficientVPR} improves VPR efficiency through scene-aware prompt tuning and adaptive local feature enhancement.

Despite these advances in representation learning, aggregation, and retrieval, large appearance discrepancies caused by seasonal changes, adverse illumination, nighttime conditions, and dynamic occlusions remain challenging for VPR. Most existing methods primarily improve how place representations are learned or matched. In contrast, AdaptVPR focuses on the complementary problem of enriching what same-place variations are observed during training by constructing challenging yet verified hard positives.

\subsection{Generative Data Augmentation}
Generative models provide an alternative way to expand the visual conditions observed during training. Early approaches mainly relied on Generative Adversarial Networks (GANs), such as CycleGAN~\cite{CycleGAN} and ToDayGAN~\cite{ToDayGAN}, to translate images across illumination, weather, or seasonal domains~\cite{porav2018adversarial,Npr}. Although such image translation can reduce specific appearance gaps, the diversity of generated conditions is often constrained by predefined source and target domains.

The emergence of diffusion models~\cite{DDPM,LDM} has enabled substantially more flexible and controllable image synthesis. Conditional street-view generation methods such as BEVControl~\cite{BEVControl} and MagicDrive~\cite{MagicDrive} exploit geometric or layout conditions to synthesize realistic driving scenes. 
DiffPlace~\cite{DiffPlace} introduces place-controllable diffusion to generate urban scenes with consistent place characteristics while varying foreground objects and weather conditions. 
Recent VPR approaches exploit synthetic domain variations differently: QdaVPR~\cite{QdaVPR} uses style transfer augmentation for adversarial domain learning, while GIFT~\cite{gift} uses structure preserving synthesis and Generative Transfer Efficacy (GTE) to guide data selection and fine tuning.
These developments demonstrate the potential of generative models to enrich place-related visual data beyond conventional image augmentation.
However, generative augmentation for VPR introduces a task-specific requirement: increasing visual diversity is useful only when the generated image remains a valid positive for the original place. A visually realistic sample may still be harmful when generation alters important architectural structures, road layouts, or other place-discriminative cues. This motivates generation strategies that jointly consider transformation difficulty and place consistency rather than relying solely on synthesis quality. AdaptVPR addresses this issue through image-dependent route scheduling, complementary geometry and diversity verification, and selective feedback refinement, enabling the construction of verified hard positives under global appearance changes, local occlusions, and their combinations.

\subsection{Controllable Generation and Agentic Image Editing}

Controllable image generation aims to modify selected visual attributes while preserving task-relevant content and structure. ControlNet~\cite{ControlNet} introduces additional spatial conditions, such as depth and edge maps, into pretrained diffusion models, while DiffEdit~\cite{Diffedit} automatically identifies regions to edit from differences between diffusion predictions. InstructPix2Pix~\cite{Instructpix2pix} further enables instruction-driven image editing directly from natural-language commands. These approaches establish important foundations for structure-aware and instruction-guided image manipulation.

Recent advances in multimodal agents further extend controllable generation toward iterative planning, tool use, and reflection. FaSTA$^*$~\cite{FaSTA*} combines high-level subtask planning with tool-path search for efficient multi-turn image editing. JarvisEvo~\cite{JarvisEvo} alternates editing, evaluation, and reflection to progressively refine image manipulation, while IMAGAgent~\cite{IMAGAgent} adopts a plan-execute-reflect framework that integrates multimodal planning, tool orchestration, and multi-expert feedback. These studies show that feedback-driven generation can improve the controllability and reliability of complex image editing, providing a relevant foundation for constructing more reliable synthetic training data.
However, applying these ideas to VPR introduces an additional requirement: generated images should exhibit sufficiently challenging visual changes while preserving the identity of the original place. Existing controllable generation and agentic editing methods are generally designed for generic visual editing objectives rather than VPR-oriented hard positive construction. AdaptVPR builds upon these developments by introducing task-aware generation and verification for constructing reliable same-place hard positives.

\section{Methodology}

\begin{figure*}[t]
    \centering
    \includegraphics[width=\textwidth]{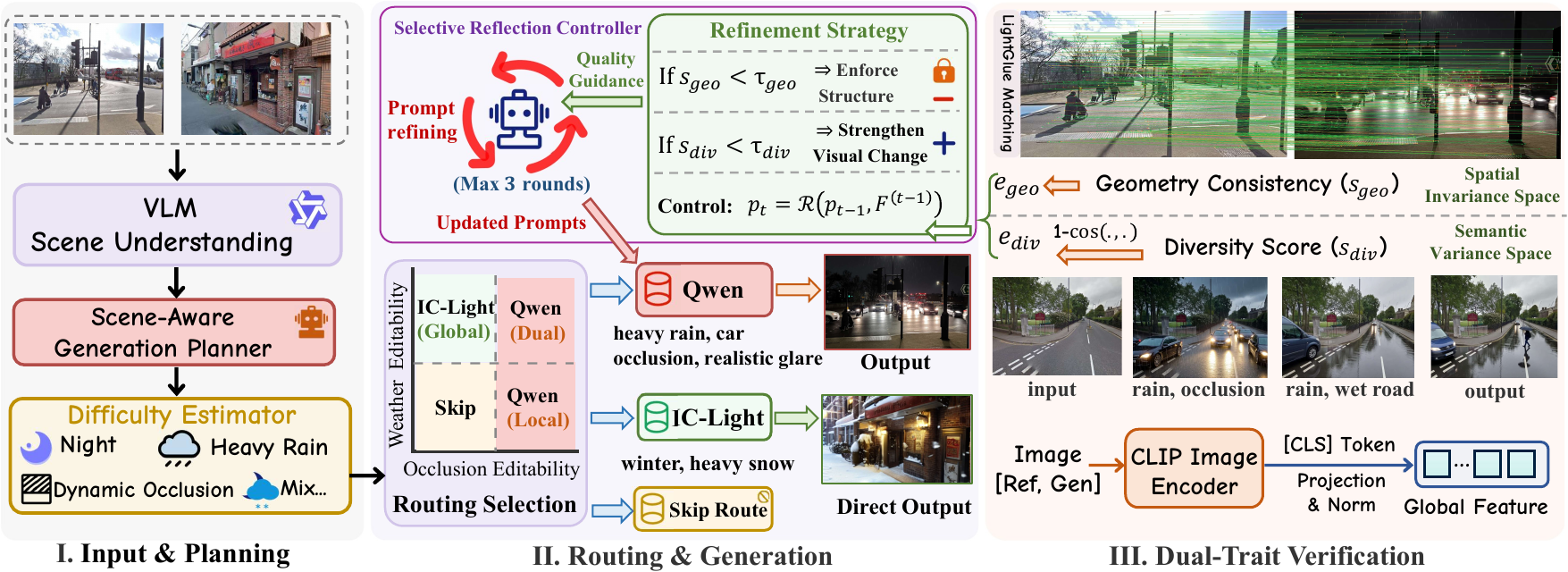}
    \caption{
    Overview of AdaptVPR. A VLM first parses each reference image and estimates its scene attributes and editability, after which a rule-based scheduler selects the Global Appearance, Local Occlusion, Dual, or Skip Route according to the parsed scene state and routing rules. The corresponding generators then synthesize candidate hard positives, which are evaluated for geometric consistency and appearance diversity. The Global Appearance Route performs a single generation followed by verification, whereas failed Local Occlusion and Dual candidates can be refined using verification feedback. Candidates satisfying the geometry and diversity gate are retained to construct AdaptCities for downstream VPR training.
    }
    \label{fig:method}
    \vspace{-0.2cm}
\end{figure*}

AdaptVPR is a training data construction framework for generating challenging same-place positives while reducing place identity drift. As illustrated in Figure~\ref{fig:method}, it integrates scene understanding, rule-based routing, route-specific generation, geometry and diversity verification, and selective prompt reflection. A VLM first analyzes scene attributes and editing feasibility, after which a rule-based scheduler determines the generation route. The Global Appearance Route modifies weather, illumination, and time of day, the Local Occlusion Route introduces dynamic occluders, and the Dual Route combines both types of changes. Generated candidates are then verified for geometric consistency and appearance diversity. Invalid Global candidates are rejected, while failed Local Occlusion and Dual candidates can be refined using verification feedback.

Section~\ref{subsec:hard_positive_generation} formulates same-place hard positive construction as a constrained generation problem. Section~\ref{subsec:planning_routing} presents scene understanding, planning, and rule-based routing. Section~\ref{subsec:geometry_diversity_verification} introduces the geometric consistency and appearance diversity criteria used to verify generated candidates. Section~\ref{subsec:self_reflective_controller} describes the selective feedback mechanism for refining failed Local Occlusion and Dual Route candidates. Section~\ref{subsec:adaptcities_construction} details the construction of AdaptCities from GSV-Cities~\cite{gsv}, while Section~\ref{subsec:vpr_training} explains how the verified hard positives are incorporated into existing VPR training.

\subsection{VPR Hard Positive Generation}
\label{subsec:hard_positive_generation}

AdaptVPR aims to construct same-place hard positives that introduce substantial visual variation while retaining sufficient structural evidence of the original place. We define a same-place hard positive as a generated image that preserves the place-defining structure of its reference image while exhibiting challenging changes in appearance or local visibility.

Let $I_{\text{ref}} \in \mathcal{D}_{\text{src}}$ denote a reference street view image from the source image pool. AdaptVPR generates a candidate $I_{\text{gen}}$ according to two complementary objectives:
\begin{itemize}
    \item \textbf{Geometric consistency.} The dominant scene layout, architectural structures, road geometry, and viewpoint should remain sufficiently consistent with $I_{\text{ref}}$ to reduce the risk of place identity drift.
    \item \textbf{Appearance diversity.} The generated image should introduce meaningful visual changes in weather, illumination, time of day, or local visibility so that it provides a challenging positive example for VPR training.
\end{itemize}

Accordingly, AdaptVPR formulates dataset expansion as a constrained generation problem rather than unconstrained image synthesis. Scene understanding and rule-based routing select an appropriate generation route, followed by route-specific synthesis and geometry-diversity verification. Failed Local Occlusion and Dual candidates can be refined within a limited reflection budget, whereas the Global Appearance Route performs only one generation followed by verification.

\subsection{Scene Planning and Rule-Based Routing}
\label{subsec:planning_routing}

Urban street view images vary substantially in scene layout, visible road area, illumination, and the feasibility of introducing local occlusions. Applying a fixed editing strategy to all images may therefore lead to implausible modifications or unnecessary structural drift. AdaptVPR addresses this issue through scene understanding and planning followed by deterministic rule-based routing.

\subsubsection{Scene Understanding and Planning}

For each reference image, the planner invokes Qwen3-VL-4B-Instruct~\cite{qwen3} to produce a structured capability dictionary containing three core fields: the weather editability score $s_{\text{weather}}$, the occlusion editability score $s_{\text{occ}}$, and the unsuitable image flag $b_{\text{bad}}$. The score $s_{\text{weather}}$ estimates whether the scene is suitable for global changes in weather, illumination, and time of day, while $s_{\text{occ}}$ estimates whether the scene contains a semantically plausible region for inserting vehicles or pedestrians while maintaining perspective consistency. The flag $b_{\text{bad}}$ identifies images unsuitable for generation, such as close-up facade fragments, walls, signs, or scenes without sufficient road context. The planner also provides candidate target conditions, occluder types and positions, and prompt constraints for subsequent generation.

Formally, scene parsing is expressed as
\begin{equation}
S = \text{VLM}(I_{\text{ref}} \mid \mathcal{M}_{\text{schema}}),
\end{equation}
where $S$ denotes the structured scene state and $\mathcal{M}_{\text{schema}}$ constrains the output format to reduce parsing failures and missing fields.

\subsubsection{Rule-Based Routing}

Based on $S$, the scheduler first determines route eligibility. The Global Appearance Route requires $s_{\text{weather}}$ to exceed its corresponding threshold, the Local Occlusion Route requires $s_{\text{occ}}$ to exceed the occlusion threshold, and the Dual Route requires both conditions to be satisfied. If $b_{\text{bad}}$ is true or no executable route is eligible, the sample is assigned to the Skip Route. When only one route is eligible, it is selected directly.

When multiple routes are eligible, the scheduler jointly considers route capability and the current quota state. The scheduling score is defined as
\begin{equation}
S_r = w_r\Delta_r + \lambda c_r,
\qquad
\Delta_r = \rho_r(N+1)-n_r ,
\end{equation}
where $\rho_r$ denotes the target ratio of route $r$, $n_r$ is the number of samples already assigned to that route, and $N$ is the total number of samples planned so far. The term $\Delta_r$ measures the quota deficit, while $\lambda$ controls the contribution of the capability score $c_r$. We define $c_r$ as $s_{\text{weather}}$ for the Global Appearance Route, $s_{\text{occ}}$ for the Local Occlusion Route, and $\min(s_{\text{weather}},s_{\text{occ}})$ for the Dual Route.

The quota state only affects selection among routes that have already satisfied their execution conditions. If the current ratio of a route falls below its minimum ratio, its deficit weight $w_r$ is increased; if the ratio exceeds its maximum ratio, that route is preferentially removed from the candidate set. The scheduler then selects the remaining route with the highest $S_r$. Thus, editability determines whether a route is suitable for the current image, while quota control prevents the generated samples from becoming overly concentrated in a single route.

The final routing decision is given by
\begin{equation}
r \in \{\text{Global Appearance},\ \text{Local Occlusion},\ \text{Dual},\ \text{Skip}\}.
\end{equation}
For $r \neq \text{Skip}$, the planned scene attributes are converted into a generation command containing the target condition, occluder type, insertion position, and initial prompt. The resulting command is then passed to the corresponding generation route.
\begin{figure}[t]
\centering
\setlength{\fboxsep}{0pt}
\setlength{\fboxrule}{0.6pt}

\fcolorbox{gray!65}{white}{
\begin{minipage}{0.98\columnwidth}

\colorbox{gray!65}{
\parbox{\dimexpr\linewidth-2\fboxrule\relax}{
\vspace{0.30em}
\hspace{0.9em}
\textcolor{white}{\textbf{Example of AdaptVPR Structured Agent Record}}
\vspace{0.30em}
}}

\vspace{0.65em}

\hspace{1.0em}
\begin{minipage}{0.90\linewidth}
\scriptsize\ttfamily

\{\\
\hspace*{1em}\textcolor{blue!70!black}{``agent\_input''}: \{\\
\hspace*{2em}\textcolor{black}{``file\_name''}: ``reference.jpg'',\\
\hspace*{2em}\textcolor{black}{``city''}: ``London'',\\
\hspace*{2em}\textcolor{black}{``round''}: 1\\
\hspace*{1em}\},\\[0.12em]

\hspace*{1em}\textcolor{green!55!black}{``scene\_state''}: \{\\
\hspace*{2em}\textcolor{black}{``weather\_score''}: 0.73,\\
\hspace*{2em}\textcolor{black}{``occlusion\_score''}: 0.61,\\
\hspace*{2em}\textcolor{black}{``street\_scene\_quality''}: ``good'',\\
\hspace*{2em}\textcolor{black}{``road\_visibility''}: ``clear'',\\
\hspace*{2em}\textcolor{black}{``sky\_visibility''}: ``partial'',\\
\hspace*{2em}\textcolor{black}{``global\_weather\_risk''}: ``low''\\
\hspace*{1em}\},\\[0.12em]

\hspace*{1em}\textcolor{orange!85!black}{``agent\_action''}: \{\\
\hspace*{2em}\textcolor{black}{``route''}: ``dual'',\\
\hspace*{2em}\textcolor{black}{``selected\_model''}: ``Qwen-LightX2V'',\\
\hspace*{2em}\textcolor{black}{``weather''}: ``rain'',\\
\hspace*{2em}\textcolor{black}{``occlusion''}: ``vehicle'',\\
\hspace*{2em}\textcolor{black}{``position''}: ``foreground road lane'',\\
\hspace*{2em}\textcolor{black}{``prompt''}: ``preserve road layout;\\
\hspace*{5em}add rain and one legal vehicle occluder''\\
\hspace*{1em}\},\\[0.12em]

\hspace*{1em}\textcolor{violet!75!black}{``verification\_feedback''}: \{\\
\hspace*{2em}\textcolor{black}{``s\_geo''}: 0.81,\\
\hspace*{2em}\textcolor{black}{``s\_div''}: 0.22,\\
\hspace*{2em}\textcolor{black}{``passed''}: true\\
\hspace*{1em}\}\\
\}\\

\end{minipage}

\vspace{0.65em}
\end{minipage}
}

\caption{
Example of the structured agent record used in AdaptVPR. Scene diagnostics and the scheduled generation action are recorded during planning and routing, while verification feedback is appended after candidate generation to determine acceptance or guide prompt refinement for Local Occlusion and Dual candidates.
}
\label{fig:planning_json}
\vspace{-0.2cm}
\end{figure}

Figure~\ref{fig:planning_json} illustrates the structured record used to connect planning, generation, and verification. The scene-state and action fields are populated during planning and routing, respectively, while the verification-feedback field is appended after generation and reused for prompt refinement when the Local Occlusion or Dual Route enters the reflection loop.

\subsubsection{Route-Aware Generation}

AdaptVPR implements three executable generation routes using two complementary generator backends. The Global Appearance Route uses IC-Light~\cite{iclight} for scene-level appearance transformation, whereas the Local Occlusion Route and Dual Route use Qwen-LightX2V~\cite{qwen3vl,lightx2v} with route-conditioned prompts.
The system adaptively dispatches each sample to a dedicated generation branch according to the routing decision $r$. Figure~\ref{fig:method} illustrates this route-conditioned dispatch, where IC-Light handles the Global Appearance Route and Qwen-LightX2V handles the Local Occlusion and Dual Routes.

\textbf{Global Appearance Route.}
For global changes in weather, illumination, and time of day, AdaptVPR uses IC-Light as the generation backend. The generation prompt, whose structured representation is exemplified in Figure~\ref{fig:planning_json}, specifies the target visual condition while explicitly constraining the original viewpoint, road layout, and architectural structure. This route is intended to alter image-wide appearance while retaining the dominant spatial configuration of the reference scene. Each Global Appearance candidate is generated once and then subjected to verification and fast rejection.

\textbf{Local Occlusion and Dual Routes.}
For localized modifications or compound visual changes, AdaptVPR uses Qwen-LightX2V with route-conditioned prompts. The Local Occlusion Route introduces realistic dynamic occluders, such as vehicles or pedestrians, into plausible ground-plane regions while constraining the original viewpoint and scene layout. For original scenes with dense vehicle presence, this route also supports inverse de-cluttering edits by removing a small number of vehicles and naturally completing the vacated regions with context-consistent background content.The Dual Route combines such local occlusions with global changes in weather, illumination, or time of day. The two editing factors are specified through separate prompt fields, allowing their constraints to be refined independently at the instruction level. If a Local Occlusion or Dual candidate fails verification, the route and generator remain fixed while the prompt can be refined within the reflection budget.
Samples assigned to the Skip Route do not enter the generation stage.

\subsection{Geometry and Diversity Verification Scheme}
\label{subsec:geometry_diversity_verification}

As illustrated in Figure~\ref{fig:method}, each synthesized candidate $I_{\text{gen}}$ is evaluated using complementary geometric and appearance criteria. Since automatic metrics cannot fully certify geographical identity, the verifier is designed as a proxy mechanism for reducing structural drift while ensuring that the generated image remains sufficiently challenging for VPR training.

\subsubsection{Geometric Consistency Measurement}

The fundamental requirement of VPR data augmentation is to reduce the risk of place-identity drift while introducing appearance changes. Since exact place identity cannot be fully certified automatically, we use geometric consistency as a proxy verification signal. Specifically, we adopt a local feature extraction and matching pipeline composed of SuperPoint~\cite{detone2018superpoint} and LightGlue~\cite{lindenberger2023lightglue}\footnote{We use the SuperPoint and LightGlue implementation released with EarthMatch~\cite{Berton_2024_EarthMatch} through the VisMatch: \url{https://github.com/gmberton/vismatch}.}. Let $E_{\text{ref}}$ and $E_{\text{gen}}$ denote the keypoints and associated descriptors extracted from $I_{\text{ref}}$ and $I_{\text{gen}}$, respectively. LightGlue establishes a robust match set $\mathcal{M}$.

To filter spatial distortions or semantic drift, such as altered building boundaries or warped road topologies, we use RANSAC to estimate a homography matrix $H$. The geometric consistency score $s_{\text{geo}}$ is defined as the final inlier ratio:
\begin{equation}
s_{\text{geo}} =
\frac{
\sum_{m \in \mathcal{M}}
\mathbb{I}\left[
\text{dist}\left(m_{\text{gen}}, H m_{\text{ref}}\right) < \epsilon
\right]
}{
|\mathcal{M}|
},
\end{equation}
where $\text{dist}(\cdot)$ denotes the forward transfer error, and $\epsilon$ is a preset pixel threshold. A higher $s_{\text{geo}}$ suggests stronger geometric consistency between the reference and generated images, but it should be interpreted as a proxy for reducing structural drift rather than a complete proof of place-identity preservation. For the Local Occlusion Route and Dual Route, newly inserted occluders may naturally reduce reliable correspondences in the edited region; therefore, the geometric proxy is interpreted together with the route objective as part of the verification design under route-specific editing conditions.

\subsubsection{Appearance Diversity Measurement}

An effective hard positive should also introduce sufficient perceptual domain shift to avoid trivial representations during training. We define the appearance diversity score as the CLIP feature distance between the reference image and the generated image. Let $\Phi_{\text{CLIP}}(\cdot)$ denote the visual encoder of a pretrained CLIP model:
\begin{equation}
s_{\text{div}} =
1 -
\frac{
\Phi_{\text{CLIP}}(I_{\text{ref}}) \cdot
\Phi_{\text{CLIP}}(I_{\text{gen}})
}{
\|\Phi_{\text{CLIP}}(I_{\text{ref}})\|_{2}
\|\Phi_{\text{CLIP}}(I_{\text{gen}})\|_{2}
}.
\end{equation}
For the Global Appearance Route, where the intended transformation affects the entire image, $s_{\text{div}}$ reflects whether weather, illumination, and time of day have changed sufficiently. 
For the Local Occlusion Route and Dual Route, occluder visibility and placement plausibility are specified as prompt-level constraints during generation and reflection. 

\subsection{Closed-Loop Self-Reflective Controller}
\label{subsec:self_reflective_controller}

The cornerstone of AdaptVPR is a closed-loop self-reflective controller that coordinates iterative interactions among the generation branches, the geometry-diversity verification module, and the prompt rewriting mechanism. Figure~\ref{fig:method} highlights this asymmetric feedback mechanism: only failed Local Occlusion and Dual candidates re-enter generation through prompt updates, whereas Global Appearance candidates follow one-shot generation with fast rejection. Unlike an open-loop synthesis pipeline that executes a static instruction only once, the controller continuously refines subsequent generation instructions according to verification feedback. Specifically, the Local Occlusion Route and Dual Route support iterative self-reflection: when a candidate fails route-specific verification, the controller converts the verifier feedback into a refined prompt for the next generation attempt. The Global Appearance Route follows one-shot generation with fast rejection and does not participate in this iterative process. Throughout reflection, the route and generator determined during planning remain fixed, while only the generation instruction is updated according to verification results, forming a closed feedback loop of generation, verification, and prompt refinement for reflection-enabled routes.

\textbf{Agent state and per-round observation.}
Let $p_0$ denote the initial prompt produced during planning. The generator first uses $p_0$ to synthesize the initial candidate $I_{\mathrm{gen}}^{(0)}$, which is subsequently evaluated by the geometry-diversity verifier to obtain feedback $F^{(0)}$. If the initial candidate fails verification and the selected route supports reflection, the controller starts iterative refinement rounds indexed by $t=1,\ldots,K$, where $K$ denotes the maximum reflection budget. At reflection round $t$, the agent observes
\begin{equation}
O_t =
\{I_{\mathrm{ref}}, r, g, c_w, c_o, b,
p_{t-1}, I_{\mathrm{gen}}^{(t-1)}, F^{(t-1)}\},
\end{equation}
where $I_{\mathrm{ref}}$ is the reference image, $r$ is the fixed route determined by the planner, $g=\text{Qwen-LightX2V}$ is the corresponding fixed generator~\cite{lightx2v}. Both the Local Occlusion Route and Dual Route use Qwen-LightX2V with route-conditioned prompts. The variables $c_w$ and $c_o$ denote the target weather condition and occluder type, respectively, while $b$ denotes the candidate local insertion position. The term $p_{t-1}$ is the prompt used to generate the previous candidate $I_{\mathrm{gen}}^{(t-1)}$, and $F^{(t-1)}$ denotes the corresponding verifier feedback, including the geometric score, diversity score, route-specific editing diagnostics, and their pass or fail states.

\textbf{Executable action output.}
Based on the previous generation instruction and its verification feedback, the controller produces a refined prompt according to
\begin{equation}
p_t =
\mathcal{R}\left(p_{t-1}, F^{(t-1)}\right),
\qquad t=1,\ldots,K,
\end{equation}
where $\mathcal{R}$ denotes the prompt refinement function. The refinement function updates the generation instruction according to the failure modes identified by the verifier. The executable action for the next generation attempt is therefore defined as
\begin{equation}
A_t =
\{r, g, c_w, c_o, b, p_t\}.
\end{equation}
The generator then executes $A_t$ to produce $I_{\mathrm{gen}}^{(t)}$, which is re-evaluated by the geometry-diversity verifier to obtain $F^{(t)}$. 
For the Local Occlusion Route, the refined prompt encodes language-level constraints on the inserted occluder, such as its type, approximate position, relative scale, and desired visibility. These attributes are not optimized as separate continuous control variables; instead, they are expressed as textual constraints passed to the selected Qwen-LightX2V model. For the Dual Route, the refined prompt combines global appearance changes in weather, illumination, and time of day with the local occlusion constraint, encouraging stronger appearance diversity while preserving the original place structure.

\textbf{Route-specific verification gates.}
Each generated candidate must pass a VPR-oriented  geometry-diversity gate matched to its route. The verifier evaluates two complementary properties: geometric consistency and appearance diversity. Geometric consistency measures whether the generated image preserves the original road layout, architectural structure, and camera viewpoint. Appearance diversity measures whether the candidate introduces sufficiently strong global appearance changes in weather, illumination, and time of day, or effective local occlusion changes, to serve as a hard positive.

Route-specific prompts define the intended editing content. The Global Appearance Route specifies global changes in weather, illumination, and time of day; the Local Occlusion Route specifies the occluder type, insertion position, scale, and visibility; and the Dual Route combines both types of generation constraints. These prompt-level constraints guide image synthesis, whereas automatic acceptance is determined solely by the geometric consistency score and the appearance diversity score:  
\begin{equation}
\mathcal{T}_r(I_{\mathrm{gen}}^{(t)}) =
\mathbb{I}\left[s_{\mathrm{geo}}^{(t)} \ge \tau_g^r\right]
\cdot
\mathbb{I}\left[s_{\mathrm{div}}^{(t)} \ge \tau_d^r\right],
\end{equation}
where $\tau_g^r$ and $\tau_d^r$ denote the route-specific thresholds for geometric consistency and appearance diversity, respectively.

\textbf{Prompt-level reflection.}
When a candidate from the Local Occlusion Route or Dual Route fails verification, the agent converts the feedback into prompt-level repair instructions rather than restarting from an unconstrained plan. The reflection result then guides the next round of prompt refinement. If the diversity score is insufficient, the generated image is too similar to the reference image and is not challenging enough. The agent then strengthens the prompt by asking for more visible rain, fog, wet road surfaces, or a more salient vehicle occluder closer to the camera. Conversely, if the geometric consistency score decreases, the edit is considered too aggressive and may have damaged road topology, building outlines, lane markings, or viewpoint. The agent then weakens the editing tendency and adds stronger structural anchors, such as preserving the exact road layout, maintaining the original camera perspective, and keeping building geometry and lane markings unchanged. Through this feedback-driven rewriting, the agent balances appearance variation against place-identity preservation across rounds.

\textbf{Stopping and trajectory recording.}
For the Local Occlusion Route and Dual Route, we index the initial generation as $t=0$ and the subsequent feedback-guided prompt-refinement attempts as $t=1,\ldots,K$. In our implementation, $K=3$ denotes three reflection attempts after the initial generation; therefore, each sample from the Local Occlusion Route or Dual Route can be generated at most $K+1$ times. After each generation round, the candidate is evaluated by the geometry-diversity gate, and the loop terminates once the candidate passes the gate or the reflection budget is exhausted:
\begin{equation}
\mathrm{Stop}_t =
\mathcal{T}_r(I_{\mathrm{gen}}^{(t)}) \lor (t=K).
\end{equation}
The Global Appearance Route is not included in this loop: it follows one-shot generation followed by verification and fast rejection. The Skip Route does not invoke image generation. If a candidate from the Local Occlusion Route or Dual Route passes before the maximum reflection round, it is retained as a valid hard positive; otherwise, it is excluded from the final training set. For traceability, the system stores the full trajectory as a structured reflection record, including the round index, prompt, generated image path, geometric and diversity scores, diagnostic feedback, final output path, and final success or failure flag. This design makes the generation process both automatically controlled and interpretable.

\subsection{AdaptCities Dataset Construction}
\label{subsec:adaptcities_construction}

To instantiate AdaptVPR at scale, we apply the generation pipeline to GSV-Cities, which serves as the source image pool for constructing AdaptCities. The resulting dataset contains verified synthetic same-place hard positives covering global appearance changes, local occlusions, and their combinations. Dataset construction is organized into four automated stages for scalable and reproducible processing.

\textbf{Step 1: deterministic sampling and semantic parsing.} To ensure balanced geographical coverage, we perform fixed-random-seed quota sampling across the 23 metropolitan areas in GSV-Cities. Each image is parsed by the VLM-based semantic analyzer to generate a localized physical attribute dictionary, which provides state tokens for downstream routing and generation operations.

\textbf{Step 2: route scheduling and planning-time risk control.}
The state tokens are converted into structured generation prompts and passed to the rule-based scheduler described above. Using $s_{\text{weather}}$, $s_{\text{occ}}$, and $b_{\text{bad}}$, the scheduler determines the eligible generation routes. Images identified as unsuitable or having no eligible route are assigned to the Skip Route. When multiple routes are eligible, the quota-aware score $S_r$ determines the final route $r$. Thus, the editability scores determine route eligibility, while the scene attributes support subsequent risk control.

Following initial routing, a planning-time review checks whether the selected transformation satisfies the scene constraints. For $r\in\{\text{Local Occlusion},\text{Dual}\}$, the planned occluder type, insertion position, relative scale, and visibility are encoded in the initial prompt $p_0$. If local insertion is high risk but global editing remains reliable, the sample is reassigned to the Global Appearance Route. For this route, candidate editing conditions are restricted according to road and sky visibility, vegetation density, and estimated weather risk, thereby avoiding overly aggressive transformations in unsuitable scenes. The VLM rationale provides only auxiliary context for prompt construction and does not override the rule-based decision $r$.

This review precedes image synthesis and is distinct from the post-generation gate $\mathcal{T}_r$ applied to each generated candidate. Once $r$ is confirmed, the route and generator $g$ remain fixed throughout subsequent synthesis and reflection; only $p_t$ is updated according to verifier feedback $F^{(t-1)}$.

\textbf{Step 3: route-driven iterative synthesis.} Global appearance transformations in weather, illumination, and time of day are assigned to IC-Light and follow one-shot generation with fast rejection, while candidates from the Local Occlusion Route and Dual Route are synthesized by Qwen-LightX2V and may enter feedback-guided prompt refinement after the initial generation. We index the initial generation as round $0$ and allow at most $K=3$ additional reflection attempts for the Local Occlusion Route and Dual Route. The refinement template, illustrated in Fig.~\ref{fig:prompt_refinement_template}, takes the current prompt and diagnostic feedback as input, then returns revised directives that are combined with route-specific constraints for the next generation round.

\begin{figure}[t]
\centering
\setlength{\fboxsep}{0pt}
\setlength{\fboxrule}{0.6pt}

\fcolorbox{gray!70}{white}{
\begin{minipage}{0.96\linewidth}

\colorbox{gray!65}{
\parbox{\dimexpr\linewidth-2\fboxrule\relax}{
\vspace{0.28em}
\hspace{0.8em}
\textcolor{white}{\textbf{Prompt Refinement Template}}
\vspace{0.28em}
}}

\vspace{0.55em}

\hspace{0.8em}
\begin{minipage}{0.88\linewidth}
\footnotesize

\textbf{Role.}
You are an image-generation prompt optimization expert.

\vspace{0.3em}
\textbf{Input.}
Current prompt $p_t$ and verification feedback indicating geometry or diversity issues.

\vspace{0.35em}
\textbf{Rewrite rules.}

\hspace{0.8em}\textcolor{blue!70!black}{\textit{Geometry issue:}}
preserve the exact road layout, original perspective, building geometry, and lane markings.

\hspace{0.8em}\textcolor{orange!85!black}{\textit{Diversity issue:}}
strengthen the intended visual change or make the occluder more salient and realistic.

\vspace{0.35em}
\textbf{Output.}
Return structured failure analysis and revised generation directives; the controller combines them with fixed route-specific constraints to form the prompt for the next generation round.

\end{minipage}

\vspace{0.55em}
\end{minipage}
}

\caption{
Prompt refinement template used to convert verification feedback into the next generation instruction for the Local Occlusion and Dual Routes.
}
\label{fig:prompt_refinement_template}
\end{figure}

\textbf{Step 4: post-verification and quality filtering.} Every synthesized candidate in the full construction pipeline is filtered solely based on geometric consistency and appearance diversity. We use fixed route-specific thresholds for the three executable routes. Global editing primarily modifies weather, illumination, and time of day while being expected to preserve most scene structures; therefore, it uses a strict geometric consistency threshold with a moderate appearance diversity requirement. Local editing introduces foreground occluders while preserving the global scene appearance as much as possible, and thus uses the strictest geometric consistency threshold together with a route-specific appearance diversity threshold. The visibility and plausibility of the inserted occluder are constrained only through the generation prompt and are not used as additional post-verification criteria. Dual editing combines global appearance changes with local occlusion. Since changes in weather, illumination, and time of day, together with foreground occlusion, can reduce local feature correspondences, we adopt a relatively more tolerant geometric consistency threshold while retaining a sufficient appearance diversity requirement. For rain-plus-vehicle Dual samples, we use a predefined relaxed appearance diversity threshold to accommodate the combined appearance variation caused by weather changes and local occlusion. All thresholds are used only to filter synthesized training samples and are not tuned on downstream VPR test sets.

\begin{table}[t]
\centering
\caption{
Composition of AdaptCities by generation route and subtype.
}
\label{tab:adaptcities_statistics}

\small
\setlength{\tabcolsep}{4.5pt}
\renewcommand{\arraystretch}{1.1}

\begin{tabular}{@{}lrr@{}}
\toprule
Route / Subtype & \# Images & Share (\%) \\
\midrule

\rowcolor{totalbg}
\textbf{Total}
& 160,000 & 100.0 \\

\midrule
\rowcolor{globalbg}
\textcolor{globaltext}{\textbf{Global Appearance Route}}
& 50,239 & 31.4 \\
\quad Overcast
& 29,607 & 18.5 \\
\quad Snowy
& 9,367 & 5.9 \\
\quad Nighttime
& 5,907 & 3.7 \\
\quad Other global conditions
& 5,358 & 3.3 \\

\addlinespace[2pt]
\rowcolor{localbg}
\textcolor{localtext}{\textbf{Local Occlusion Route}}
& 45,804 & 28.6 \\
\quad Curbside / parked-vehicle occlusion
& 21,840 & 13.7 \\
\quad Road-traffic occlusion
& 13,285 & 8.3 \\
\quad Other local occlusions
& 10,679 & 6.7 \\

\addlinespace[2pt]
\rowcolor{dualbg}
\textcolor{dualtext}{\textbf{Dual Route}}
& 63,957 & 40.0 \\
\quad Rainy night + vehicle occlusion
& 24,214 & 15.1 \\
\quad Snowy + vehicle occlusion
& 21,748 & 13.6 \\
\quad Other combinations
& 17,995 & 11.2 \\

\bottomrule
\end{tabular}
\end{table}

As summarized in Table~\ref{tab:adaptcities_statistics}, after route-specific filtering, AdaptCities contains 160K verified synthetic hard positives for training, generated from 88,989 unique GSV-Cities reference images. For each verified instance $\tilde{I}_i$, the dataset inherits its parent geographical label:
\begin{equation}
(\tilde{I}_i, y_i) \in \mathcal{D}_{\text{adapt}},
\quad
y(\tilde{I}_i)=y(I_i).
\end{equation}
The final training pool is constructed by combining the original GSV-Cities training set with AdaptCities:
\begin{equation}
\mathcal{D}_{\text{total}} =
\mathcal{D}_{\text{base}} \cup \mathcal{D}_{\text{adapt}}.
\end{equation}
Each entry in $\mathcal{D}_{\text{adapt}}$ is accompanied by a metadata tracking log for subsequent auditing and analysis:
\begin{equation}
m_i =
\{r_i, c_w, c_o, p_i, s_{\text{geo}}, s_{\text{div}}, t, \texttt{passed}\},
\end{equation}
where $r_i$ denotes the route selection, $c_w$ and $c_o$ specify weather and occlusion conditions, $p_i$ is the final prompt, and $t$ records the exact number of reflection rounds. This explicit structure supports traceable data auditing and ablation studies. Redistribution of generated images follows the license and privacy constraints of the source GSV-Cities data.

\subsection{VPR Training Protocol}
\label{subsec:vpr_training}

AdaptVPR is model-agnostic and requires no modifications to the neural backbone, feature aggregator, or loss function of existing VPR baselines. By expanding the local positive sample manifold, our framework can be integrated into standard metric learning objectives.

For a query anchor $I_q$, its corresponding geographical positive set $P_q$ and negative set $N_q$ within a training batch $B$ are defined as
\begin{equation}
P_q = \{p \mid y_p=y_q,\, p \neq q\},
\quad
N_q = \{n \mid y_n \neq y_q\}.
\end{equation}
Thanks to the proxy-verified place consistency of AdaptCities, verified synthetic instances directly expand the cardinality of $P_q$ and introduce explicitly constructed hard positive variations into each mini-batch.

During optimization, each VPR baseline retains its original metric-learning objective. For baselines trained with Multi-Similarity Loss~\cite{mixvpr,boq}, the objective can be written as
\begin{equation}
\begin{aligned}
&\mathcal{L}_{\text{MS}} =
\frac{1}{|B|}
\sum_{q \in B}
\left\{
\frac{1}{\alpha}
\ln\left[
1 + \sum_{p \in P_q}
e^{-\alpha(\mathcal{S}_{qp}-\lambda)}
\right] \right. 
\\
&\quad + \left. 
\frac{1}{\beta}
\ln\left[
1 + \sum_{n \in N_q}
e^{\beta(\mathcal{S}_{qn}-\lambda)}
\right]
\right\},
\end{aligned}
\end{equation}
where $\mathcal{S}$ denotes the pairwise similarity matrix, and $\alpha$, $\beta$, and $\lambda$ are hyperparameters controlling pair weights.

\section{Experiments}
\subsection{Datasets and Evaluation Metrics}

We evaluate AdaptVPR on ten benchmark settings covering large-scale urban retrieval, viewpoint variation, seasonal changes, adverse illumination, and partial occlusion\footnote{Dataset download utilities are provided by the VPR Datasets Downloader~\cite{benchmark}: \url{https://github.com/gmberton/VPR-datasets-downloader}.}. 
The statistics and primary visual variations of these datasets are summarized in Table~\ref{tab:dataset_summary}. 
\textbf{Pitts30k-test} and \textbf{Pitts250k-test}~\cite{torii2013visual} are derived from geotagged Google Street View imagery of Pittsburgh. Pitts30k-test is widely used for evaluating urban place recognition under substantial viewpoint changes, while Pitts250k-test provides a considerably larger database for large-scale retrieval evaluation.
\textbf{MSLS-val}~\cite{warburg2020mapillary} is the public validation split of the Mapillary Street-Level Sequences dataset, which contains imagery collected across 30 cities on six continents and exhibits diverse variations in viewpoint, illumination, weather, season, camera characteristics, and dynamic objects.
\textbf{Tokyo24/7}~\cite{densevlad} contains urban queries captured under substantially different viewpoints and illumination conditions, with pronounced day--night changes.
\textbf{Nordland}~\cite{sunderhauf2013we} records the same long-distance train route across four seasons using a forward-facing camera, providing a controlled benchmark for severe seasonal appearance changes with limited viewpoint variation. Following the commonly adopted protocol, we use the winter traversal as queries and the summer traversal as database. We additionally evaluate on \textbf{Nordland$\star$}~\cite{sunderhauf2013we}, where a subset of the summer sequence is used as queries against the full winter database.
\textbf{SVOX}~\cite{SVOX} evaluates cross-condition VPR under diverse weather and illumination changes, including rain, sunlight, snow, nighttime, and overcast conditions.
Finally, we use three query sets from \textbf{SF-XL}~\cite{cosplace,barbarani2023local}. \textbf{SF-XL-v1} contains Flickr queries exhibiting viewpoint, camera, illumination, and scene-content variations; \textbf{SF-XL-Occlusion} focuses on severe foreground occlusions, mainly caused by vehicles and crowds; and \textbf{SF-XL-Night} evaluates place recognition under challenging nighttime illumination together with viewpoint changes.

\begin{table}[t]
\caption{Statistics and dominant visual variations of the benchmark datasets used for evaluation. The three SF-XL query sets share the same large-scale database.}
\label{tab:dataset_summary}

\small
\centering
\begingroup
\setlength{\tabcolsep}{1.8pt}
\renewcommand{\arraystretch}{1.12}

\begin{tabular}{@{}c l l r r@{}}
\toprule
\rowcolor{headerbg}
ID & Dataset & Primary Variation & Database & Queries \\
\midrule

1 & Pitts30k-test
& Viewpoint
& 10,000 & 6,816 \\

2 & MSLS-val
& Mixed conditions
& 18,871 & 740 \\

3 & Tokyo24/7
& Day--night
& 75,984 & 315 \\

4 & Nordland
& Seasonal
& 27,592 & 27,592 \\

5 & SVOX
& Weather / illumination
& 17,166 & 14,278 \\

6 & Pitts250k-test
& Large-scale urban
& 83,952 & 8,280 \\

7 & Nordland$\star$
& Cross-season
& 27,592 & 2,760 \\

\addlinespace[2pt]
\rowcolor{sfxlbg}
8 & SF-XL-v1
& Viewpoint / appearance
&
& 1,000 \\

\rowcolor{sfxlbg}
9 & SF-XL-Occlusion
& Occlusion
&
& 76 \\

\rowcolor{sfxlbg}
10 & SF-XL-Night
& Nighttime
& \multirow{-3}{*}{2,805,840}
& 466 \\

\bottomrule
\end{tabular}
\endgroup
\vspace{-0.2cm}
\end{table}

We report Recall@N (R@N) as the primary evaluation metric, defined as the proportion of query images for which at least one of the top-$N$ retrieved database images is considered a correct match. Following standard VPR evaluation protocols~\cite{benchmark,gsv,cosplace}, a prediction is regarded as correct within 25\,m for Pitts30k, Pitts250k, Tokyo24/7, SVOX, and the SF-XL benchmarks, while MSLS additionally requires the heading difference to be within $40^\circ$. For the sequence-based Nordland benchmark, correctness is determined by temporal alignment, using a tolerance of $\pm 10$ frames for Nordland and a stricter one-frame tolerance for Nordland$\star$~\cite{boq}.

\subsection{Implementation Details}

For fair comparison, each VPR baseline retains its original backbone, feature aggregation module, optimization strategy, and loss function, with AdaptVPR-generated same-place hard positives added to the training data. 
We select the checkpoint with the highest R@1 on Pitts30k-val for evaluation on the remaining test datasets. Following standard VPR evaluation protocols\footnote{We organize the evaluation following the Deep Visual Geo-Localization Benchmark~\cite{benchmark} and VPR-methods-evaluation from EigenPlaces~\cite{eigenplaces}: \url{https://github.com/gmberton/deep-visual-geo-localization-benchmark}.}~\cite{benchmark,boq,patchvlad}, performance is reported using Recall@N, and the baseline settings are described below.

To perform quality filtering for generated samples, we set fixed verification thresholds for each generation route. For the Global, Local, and Dual routes, the geometric consistency threshold $\tau_{\rm geo}$ is set to $0.78$, $0.82$, and $0.72$, respectively, while the appearance diversity threshold $\tau_{\rm div}$ is set to $0.15$, $0.09$, and $0.20$. For rainy-weather-plus-vehicle Dual samples, $\tau_{\rm div}$ is set to $0.12$, while $\tau_{\rm geo}$ remains $0.72$. Occluder visibility and placement plausibility in the Local and Dual routes are used only as prompt-level constraints during generation and prompt refinement and are not included in post-verification. Sample acceptance is determined solely by $s_{\rm geo}$ and $s_{\rm div}$ under the corresponding thresholds. All thresholds are used only for quality filtering of generated training samples and remain fixed across downstream VPR evaluations.
All VLM planning, image generation, and VPR experiments were conducted on a single NVIDIA RTX PRO 6000 GPU.

\textbf{BoQ:}
We use the official implementation of BoQ~\cite{boq} with a DINOv2-B backbone. The model is trained with the AdamW optimizer using an initial learning rate of $1\times10^{-4}$, a batch size of 128. The training input size is set to $280\times280$, the evaluation input size is set to $322\times322$, and the maximum number of training epochs is set to 40. 
For the AdaptVPR variant, we add hard positives generated by AdaptVPR to the training data while keeping the original BoQ feature aggregation structure and loss function unchanged.

\textbf{SALAD:}
We use the official implementation of SALAD~\cite{salad}, while keeping the original training paradigm unchanged and only introducing AdaptVPR-generated same-place hard positives into the training data. SALAD produces 8448-dimensional global descriptors. The image resolution is 224$\times$224 for training and 322$\times$322 for inference, respectively, and training is conducted for 4 epochs.

\textbf{EDTformer:}
We use the official implementation of EDTformer~\cite{EDTformer}, and incorporate AdaptVPR-generated same-place hard positives into training under the same training framework as the original method. The global descriptor dimension of this method is 4096. The image resolution is 224$\times$224 for training and 322$\times$322 for inference, and the maximum number of training epochs is set to 15. Its original backbone, feature aggregation module, optimization strategy, and loss function are kept unchanged.

\textbf{ImAge:}
We use the official implementation of ImAge~\cite{image} with DINOv2-B-register as the backbone. The first 8 Transformer blocks are frozen, and 8 learnable aggregation tokens and 4 register tokens are employed. AdaptVPR-generated same-place hard positives are introduced under the same training framework as the original method. The image resolution is 224$\times$224 for training and 322$\times$322 for inference, and the model produces 6144-dimensional global descriptors. The maximum number of training epochs is set to 20. The remaining backbone and feature aggregation structures are kept unchanged from the original implementation.

\subsection{Comparative Results and Analysis}

\begin{table*}[t]
\caption{
Main comparison with representative VPR methods on multiple benchmark datasets. 
All models are evaluated at a resolution of $322 \times 322$. 
Baseline results are taken from the original papers when available; otherwise, we evaluate the officially released pretrained models under the corresponding benchmark protocols.
``+AdaptVPR'' denotes training the corresponding method with our generated hard positives while keeping its model configuration unchanged.
Red and green values in parentheses denote the absolute R@1 gains and drops relative to the corresponding baseline, respectively.
}
\small
\centering
\begingroup
\setlength{\tabcolsep}{0.15mm}
\renewcommand{\arraystretch}{1.1}
\begin{tabular}{@{}l|c||ccc||ccc||ccc||ccc||ccc@{}}
\toprule
\multirow{2}{*}{Method} & \multirow{2}{*}{Dim}
& \multicolumn{3}{c||}{Pitts30k}
& \multicolumn{3}{c||}{MSLS-val}
& \multicolumn{3}{c||}{Tokyo24/7}
& \multicolumn{3}{c||}{Nordland}
& \multicolumn{3}{c}{SVOX} \\
\cline{3-17}
&
& \scalebox{0.92}{R@1} & \scalebox{0.92}{R@5} & \scalebox{0.92}{R@10}
& \scalebox{0.92}{R@1} & \scalebox{0.92}{R@5} & \scalebox{0.92}{R@10}
& \scalebox{0.92}{R@1} & \scalebox{0.92}{R@5} & \scalebox{0.92}{R@10}
& \scalebox{0.92}{R@1} & \scalebox{0.92}{R@5} & \scalebox{0.92}{R@10}
& \scalebox{0.92}{R@1} & \scalebox{0.92}{R@5} & \scalebox{0.92}{R@10} \\
\hline
CosPlace \cite{cosplace}~$_{\textcolor{blue}{\text{CVPR' 2022}}}$ & 512
& 88.4 & 94.5 & 95.7
& 82.8 & 89.7 & 92.0
& 81.9 & 90.2 & 92.7
& 58.5 & 73.7 & 79.4
& 95.4 & 97.5 & 98.1 \\
MixVPR \cite{mixvpr}~$_{\textcolor{blue}{\text{WACV' 2023}}}$ & 4096
& 91.5 & 95.5 & 96.3
& 88.0 & 92.7 & 94.6
& 85.1 & 91.7 & 94.3
& 76.2 & 86.9 & 90.3
& 97.8 & 98.9 & 99.1 \\
EigenPlaces \cite{eigenplaces}~$_{\textcolor{blue}{\text{ICCV' 2023}}}$ & 2048
& 92.5 & 96.8 & 97.6
& 89.1 & 93.8 & 95.0
& 93.0 & 96.2 & 97.5
& 71.2 & 83.8 & 88.1
& 98.0 & 99.0 & 99.2 \\
SelaVPR \cite{selavpr}~$_{\textcolor{blue}{\text{ICLR' 2024}}}$ & /
& 92.8 & 96.8 & 97.7
& 90.8 & 96.4 & 97.2
& 94.0 & 96.8 & 97.5
& 87.3 & 93.8 & 95.6
& 97.2 & 98.7 & 99.0 \\
FoL-B \cite{FoL}~$_{\textcolor{blue}{\text{AAAI' 2025}}}$ & /
& 93.1 & 96.9 & 97.7
& 91.5 & 96.2 & 96.8
& 97.5 & 98.1 & 98.4
& 85.4 & 92.7 & 94.8
& 98.4 & 99.3 & 99.5 \\
FoL-L \cite{FoL}~$_{\textcolor{blue}{\text{AAAI' 2025}}}$ & /
& 93.9 & 96.9 & 98.1
& 90.1 & 95.7 & 96.9
& 97.1 & 97.8 & 98.7
& 87.9 & 94.8 & 96.6
& 98.7 & 99.5 & 99.7 \\

\hline
SALAD \cite{salad}~$_{\textcolor{blue}{\text{CVPR' 2024}}}$ & 8448
& 92.5 & 96.4 & 97.5
& 92.2 & 96.4 & 97.0
& 94.6 & 97.5 & 97.8
& 89.7 & 95.5 & 97.0
& 98.2 & 99.3 & 99.4 \\

\rowcolor{lightshade}
\quad +AdaptVPR & 8448
& 93.2{\scriptsize\textcolor{red}{(+0.7)}} & 96.8 & 97.8
& 93.5{\scriptsize\textcolor{red}{(+1.3)}} & 96.8 & 97.2
& 94.9{\scriptsize\textcolor{red}{(+0.3)}} & 98.1 & 98.4
& 90.2{\scriptsize\textcolor{red}{(+0.5)}} & 96.2 & 97.5
& 98.6{\scriptsize\textcolor{red}{(+0.4)}} & 99.4 & 99.6 \\
BoQ \cite{boq}~$_{\textcolor{blue}{\text{CVPR' 2024}}}$ & 12288
& 93.7 & 97.1 & 97.9
& 93.8 & 96.8 & 97.0
& 96.5 & 97.8 & 98.4
& 90.6 & 96.0 & 97.5
& 98.8 & 99.4 & 99.5 \\

\rowcolor{lightshade}
\quad +AdaptVPR & 12288
& 93.8{\scriptsize\textcolor{red}{(+0.1)}} & 97.4 & 98.3
& 94.3{\scriptsize\textcolor{red}{(+0.5)}} & 97.0 & 97.8
& 96.8{\scriptsize\textcolor{red}{(+0.3)}} & 97.8 & 98.1
& 95.9{\scriptsize\textcolor{red}{(+5.3)}} & 98.5 & 99.1
& 98.9{\scriptsize\textcolor{red}{(+0.1)}} & 99.6 & 99.7 \\

EDTformer \cite{EDTformer}~$_{\textcolor{blue}{\text{TCSVT' 2025}}}$ & 4096
& 93.4 & 97.0 & 97.9
& 92.0 & 96.6 & 97.2
& 97.1 & 98.1 & 98.4
& 88.3 & 95.3 & 97.0
& 98.5 & 99.4 & 99.6 \\

\rowcolor{lightshade}
\quad +AdaptVPR & 4096
& 93.9{\scriptsize\textcolor{red}{(+0.5)}} & 97.3 & 98.2
& 93.0{\scriptsize\textcolor{red}{(+1.0)}} & 96.8 & 97.3
& 97.8{\scriptsize\textcolor{red}{(+0.7)}} & 99.1 & 99.4
& 89.5{\scriptsize\textcolor{red}{(+1.2)}} & 96.0 & 97.5
& 98.9{\scriptsize\textcolor{red}{(+0.4)}} & 99.5 & 99.6 \\
ImAge \cite{image}~$_{\textcolor{blue}{\text{NeurIPS' 2025}}}$ & 6144
& 94.0 & 97.2 & 98.0
& 93.0 & 97.0 & 97.2
& 96.2 & 98.1 & 98.4
& 93.2 & 97.6 & 98.6
& 98.5 & 99.4 & 99.6 \\

\rowcolor{lightshade}
\quad +AdaptVPR & 6144
& 94.3{\scriptsize\textcolor{red}{(+0.3)}} & 97.3 & 98.1
& 94.3{\scriptsize\textcolor{red}{(+1.3)}} & 97.4 & 98.0
& 97.1{\scriptsize\textcolor{red}{(+0.9)}} & 98.4 & 98.4
& 95.0{\scriptsize\textcolor{red}{(+1.8)}} & 98.3 & 99.0
& 98.7{\scriptsize\textcolor{red}{(+0.2)}} & 99.4 & 99.6 \\
\bottomrule
\end{tabular}
\endgroup
\vspace{-0.3cm}
\label{tab:compare_SOTA}
\end{table*}

\begin{table*}[t]
\caption{Comparison under challenging domain conditions including large-scale retrieval, cross-season changes, occlusion, and nighttime appearance variations.}
\centering
\resizebox{\textwidth}{!}{%
\begingroup
\scriptsize
\setlength{\tabcolsep}{0.3mm}
\renewcommand{\arraystretch}{1.1}
\begin{tabular}{@{}l|c||ccc||ccc||ccc||ccc||ccc@{}}
\toprule
\multirow{3}{*}{Method} & \multirow{3}{*}{Dim}
& \multicolumn{3}{c||}{\multirow{2}{*}{Pitts250k-test}}
& \multicolumn{3}{c||}{\multirow{2}{*}{Nordland$\star$}}
& \multicolumn{9}{c}{SF-XL} \\
\cline{9-17}
&
& \multicolumn{3}{c||}{}
& \multicolumn{3}{c||}{}
& \multicolumn{3}{c||}{SF-XL-v1}
& \multicolumn{3}{c||}{Occlusion}
& \multicolumn{3}{c}{Night} \\
\cline{3-17}
&
& \scalebox{0.92}{R@1} & \scalebox{0.92}{R@5} & \scalebox{0.92}{R@10}
& \scalebox{0.92}{R@1} & \scalebox{0.92}{R@5} & \scalebox{0.92}{R@10}
& \scalebox{0.92}{R@1} & \scalebox{0.92}{R@5} & \scalebox{0.92}{R@10}
& \scalebox{0.92}{R@1} & \scalebox{0.92}{R@5} & \scalebox{0.92}{R@10}
& \scalebox{0.92}{R@1} & \scalebox{0.92}{R@5} & \scalebox{0.92}{R@10} \\
\hline

SALAD \cite{salad}~$_{\textcolor{blue}{\text{CVPR' 2024}}}$ & 8448
& 95.1 & 98.5 & 99.1
& 76.0 & 89.2 & 92.0
& 88.6 & 93.5 & 94.4
& 51.3 & 65.8 & 68.4
& 46.6 & 59.0 & 62.2 \\

\rowcolor{lightshade}
\quad +AdaptVPR & 8448
& 95.5{\tiny\textcolor{red}{(+0.4)}} & 98.9 & 99.4
& 82.0{\tiny\textcolor{red}{(+6.0)}} & 93.2 & 95.4
& 93.0{\tiny\textcolor{red}{(+4.4)}} & 96.3 & 97.2
& 51.3{\tiny\textcolor{red}{(+0.0)}} & 65.8 & 69.7
& 49.8{\tiny\textcolor{red}{(+3.2)}} & 64.4& 69.3 \\
BoQ \cite{boq}~$_{\textcolor{blue}{\text{CVPR' 2024}}}$ & 12288
& 96.6 & 99.1 & 99.5
& 81.3 & 92.5 & 94.8
& 91.2 & 95.0 & 95.7
& 43.4 & 67.1 & 71.1
& 48.3 & 61.2 & 65.2 \\

\rowcolor{lightshade}
\quad +AdaptVPR & 12288
& 96.1{\tiny\textcolor{darkgreen}{(-0.5)}} & 98.9 & 99.4
& 89.6{\tiny\textcolor{red}{(+8.3)}} & 96.6 & 97.7
& 91.5{\tiny\textcolor{red}{(+0.3)}} & 95.1 & 95.8
& 52.6{\tiny\textcolor{red}{(+9.2)}} & 68.4 & 72.4
& 50.0{\tiny\textcolor{red}{(+1.7)}} & 64.2 & 68.5 \\
EDTformer \cite{EDTformer}~$_{\textcolor{blue}{\text{TCSVT' 2025}}}$ & 4096
& 95.9 & 98.8 & 99.3
& 73.1 & 86.7 & 90.1
& 92.9 & 95.5 & 96.3
& 51.3 & 67.1 & 71.1
& 53.6 & 65.7 & 68.9 \\

\rowcolor{lightshade}
\quad +AdaptVPR & 4096
& 96.1{\tiny\textcolor{red}{(+0.2)}} & 98.9 & 99.4
& 81.2{\tiny\textcolor{red}{(+8.1)}} & 92.6 & 95.4
& 93.9{\tiny\textcolor{red}{(+1.0)}} & 97.0 & 97.4
& 57.9{\tiny\textcolor{red}{(+6.6)}} & 67.1 & 71.1
& 56.2{\tiny\textcolor{red}{(+2.6)}} & 67.8 & 71.9 \\
ImAge \cite{image}~$_{\textcolor{blue}{\text{NeurIPS' 2025}}}$ & 6144
& 96.5 & 99.1 & 99.5
& 84.6 & 95.1 & 97.0
& 90.6 & 94.8 & 95.8
& 56.6 & 68.4 & 72.4
& 47.2 & 61.2 & 65.7 \\

\rowcolor{lightshade}
\quad +AdaptVPR & 6144
& 96.6{\tiny\textcolor{red}{(+0.1)}} & 99.1 & 99.5
& 89.2{\tiny\textcolor{red}{(+4.6)}} & 96.1 & 97.3
& 93.0{\tiny\textcolor{red}{(+2.4)}} & 96.0 & 96.9
& 63.2{\tiny\textcolor{red}{(+6.6)}} & 71.1 & 72.4
& 49.6{\tiny\textcolor{red}{(+2.4)}} & 61.6 & 66.7 \\
\bottomrule
\end{tabular}
\endgroup
}
\vspace{-0.3cm}
\label{tab:domain_condition}
\end{table*}

We compare AdaptVPR with representative VPR methods spanning different architectures and feature aggregation strategies, including CosPlace~\cite{cosplace}, MixVPR~\cite{mixvpr}, EigenPlaces~\cite{eigenplaces}, SelaVPR~\cite{selavpr}, FoL~\cite{FoL}, SALAD~\cite{salad}, BoQ~\cite{boq}, EDTformer~\cite{EDTformer}, and ImAge~\cite{image}. 
Table~\ref{tab:compare_SOTA} evaluates whether AdaptVPR can consistently improve strong VPR baselines under standard benchmark settings, while Table~\ref{tab:domain_condition} further examines its robustness under challenging seasonal, nighttime, and occlusion conditions. 
In addition to comparing absolute retrieval performance, we report the performance changes obtained by introducing AdaptVPR-generated hard positives into four representative baselines.

As shown in Table~\ref{tab:compare_SOTA}, AdaptVPR yields positive R@1 gains across all 20 combinations of four baselines and five benchmarks. 
These baselines cover different VPR designs, including optimal transport based aggregation in SALAD, query driven attention aggregation in BoQ, decoder based feature aggregation with learnable queries in EDTformer, and implicit aggregation with learnable aggregation tokens in ImAge.
On Pitts30k, the R@1 gains range from 0.1\% to 0.7\%, while MSLS-val improves by 0.5\%--1.3\% and Tokyo24/7 by 0.3\%--0.9\%. More pronounced improvements are observed on Nordland, where BoQ gains 5.3\% at R@1 together with 2.5\% and 1.6\% at R@5 and R@10, while ImAge improves R@1 by 1.8\%. AdaptVPR also consistently improves R@1 on SVOX by 0.1\%--0.4\%, despite the already high baseline performance. These results indicate that the benefits of AdaptVPR are not tied to a particular aggregation or retrieval architecture, but transfer consistently across heterogeneous VPR models and benchmark conditions.

As shown in Table~\ref{tab:domain_condition}, the advantage of AdaptVPR becomes more pronounced under challenging domain conditions that stress different aspects of VPR robustness. Pitts250k-test mainly evaluates large-scale urban retrieval, Nordland$\star$ emphasizes cross-season matching, while the three SF-XL subsets respectively introduce viewpoint and appearance variation, partial occlusion, and nighttime illumination changes. On Nordland$\star$, AdaptVPR improves R@1 by 6.0\%, 8.3\%, 8.1\%, and 4.6\% for SALAD, BoQ, EDTformer, and ImAge, respectively, demonstrating strong gains under seasonal shifts. The improvement is also substantial on SF-XL-Occlusion, where BoQ gains 9.2\% in R@1 and EDTformer and ImAge both gain 6.6\%. On SF-XL-Night, all four baselines improve by 1.7\%--3.2\%, while the gains on SF-XL-v1 reach up to 4.4\%. In contrast, the changes on Pitts250k-test are relatively small, ranging from $-0.5\%$ to $+0.4\%$; the only R@1 decrease across these challenging settings occurs for BoQ, with a 0.5\% drop. Overall, the larger improvements under cross-season, nighttime, and occlusion conditions indicate that AdaptVPR is effective when retrieval requires robustness to strong appearance and visibility changes.

The improvements are consistent across heterogeneous VPR baselines rather than being concentrated on a single model. For example, SALAD gains 6.0\% on Nordland$\star$ and 3.2\% on SF-XL-Night, BoQ achieves the largest gain of 9.2\% on SF-XL-Occlusion, EDTformer improves by 8.1\% on Nordland$\star$ and 6.6\% on SF-XL-Occlusion, while ImAge gains 4.6\% and 6.6\% on the same two settings. This consistency supports the model-agnostic nature of AdaptVPR: the generated same-place hard positives complement different VPR architectures by enriching the appearance variations observed for each place during training, with the largest benefits appearing under substantial domain shifts.

\begin{table*}[t]
\caption{Ablation study of synthetic quality filtering in AdaptVPR. We evaluate the effect of geometry consistency and appearance diversity for filtering generated hard-positive samples. ``Retained Ratio'' denotes the percentage of candidates retained from the generated candidate pool after filtering. Training time is averaged over 10 epochs on a single NVIDIA RTX PRO 6000 GPU. The \colorbox{first}{1st}, \colorbox{second}{2nd}, and \colorbox{third}{3rd}-best values are highlighted.}
\vspace{-0.1cm}
\small
\centering
\begingroup
\setlength{\tabcolsep}{0.55mm}
\renewcommand{\arraystretch}{1.05}
\begin{tabular}{@{}c|cc|c|c||ccc||ccc||ccc||ccc||ccc@{}}
\toprule
\multirow{2}{*}{Baseline}
& \multicolumn{2}{c|}{Filtering}
& \multirow{2}{*}{\begin{tabular}[c]{@{}c@{}}Retained\\Ratio\end{tabular}}
& \multirow{2}{*}{\begin{tabular}[c]{@{}c@{}}Training Time\\(min/epoch)\end{tabular}}
& \multicolumn{3}{c||}{Pitts30k}
& \multicolumn{3}{c||}{MSLS-val}
& \multicolumn{3}{c||}{SF-XL-Night}
& \multicolumn{3}{c||}{SF-XL-Occ.}
& \multicolumn{3}{c}{SF-XL-v1} \\
\cline{2-3}\cline{6-20}
& Geo. & Div.
&
& 
& \scalebox{0.92}{R@1} & \scalebox{0.92}{R@5} & \scalebox{0.92}{R@10}
& \scalebox{0.92}{R@1} & \scalebox{0.92}{R@5} & \scalebox{0.92}{R@10}
& \scalebox{0.92}{R@1} & \scalebox{0.92}{R@5} & \scalebox{0.92}{R@10}
& \scalebox{0.92}{R@1} & \scalebox{0.92}{R@5} & \scalebox{0.92}{R@10}
& \scalebox{0.92}{R@1} & \scalebox{0.92}{R@5} & \scalebox{0.92}{R@10} \\
\hline\multirow{4}{*}{BoQ}
& - & -
& 100.0\%
& 14.6
& \nd 93.4 & 96.7 & 97.7
& 93.1 & \nd 96.4 & \rd 96.9
& \nd 49.4 & \nd 62.2 & \rd 66.1
& \nd 50.0 & \nd 64.5 & \rd 68.4
& \nd 90.7 & \nd 94.3 & \nd 95.4 \\

& \checkmark & -
& 81.7\%
& 13.1
& \rd 93.3 & \nd 96.9 & \nd 97.8
& \nd 93.2 & 95.9 & 96.6
& \rd 49.1 & \rd 60.1 & \nd 66.5
& \rd 48.7 & 63.2 & 67.1
& \rd 89.2 & \rd 93.9 & \rd 95.2 \\

& - & \checkmark
& 83.7\%
& 13.9
& 93.2 & \rd 96.8 & \nd 97.8
& \nd 93.2 & \nd 96.4 & \nd 97.0
& 46.8 & 59.9 & 64.8
& \rd 48.7 & \nd 64.5 & \nd 69.7
& 88.6 & 93.3 & 94.5 \\

& \checkmark & \checkmark
& 65.6\%
& 13.8
& \fs 93.8 & \fs 97.4 & \fs 98.3
& \fs 94.3 & \fs 97.0 & \fs 97.8
& \fs 50.0 & \fs 64.2 & \fs 68.5
& \fs 52.6 & \fs 68.4 & \fs 72.4
& \fs 91.5 & \fs 95.1 & \fs 95.8 \\
\bottomrule
\end{tabular}
\endgroup
\vspace{-0.2cm}
\label{tab:filtering_ablation_boq_2}
\end{table*}

\begin{table*}[t]
\caption{Ablation study of different generation routes in AdaptVPR on challenging VPR benchmarks. The first row uses all unfiltered synthetic samples. Starting from the BoQ baseline, we evaluate the contribution of the Global Appearance Route, Local Occlusion Route, and Dual Route. Training time is averaged across epochs on a single NVIDIA RTX PRO 6000 GPU.}
\small
\centering
\begingroup
\setlength{\tabcolsep}{0.55mm}
\renewcommand{\arraystretch}{1.05}
\begin{tabular}{@{}c|ccc|c||ccc||ccc||ccc||ccc||ccc@{}}
\toprule
\multirow{2}{*}{Baseline}
& \multicolumn{3}{c|}{Route}
& \multirow{2}{*}{\begin{tabular}[c]{@{}c@{}}Training Time\\(min/epoch)\end{tabular}}
& \multicolumn{3}{c||}{Pitts30k}
& \multicolumn{3}{c||}{MSLS-val}
& \multicolumn{3}{c||}{SF-XL-Night}
& \multicolumn{3}{c||}{SF-XL-Occ.}
& \multicolumn{3}{c}{SF-XL-v1} \\
\cline{2-4}\cline{6-20}
& Global & Local & Dual
&
& \scalebox{0.92}{R@1} & \scalebox{0.92}{R@5} & \scalebox{0.92}{R@10}
& \scalebox{0.92}{R@1} & \scalebox{0.92}{R@5} & \scalebox{0.92}{R@10}
& \scalebox{0.92}{R@1} & \scalebox{0.92}{R@5} & \scalebox{0.92}{R@10}
& \scalebox{0.92}{R@1} & \scalebox{0.92}{R@5} & \scalebox{0.92}{R@10}
& \scalebox{0.92}{R@1} & \scalebox{0.92}{R@5} & \scalebox{0.92}{R@10} \\
\hline
\multirow{5}{*}{BoQ}
& \multicolumn{3}{c|}{Unfiltered}
& 14.6
& \nd 93.4 & \rd 96.7 & \rd 97.7
& \rd 93.1 & \nd 96.4 & \nd 96.9
& \rd 49.4 & \nd 62.2 & \nd 66.1
& \rd 50.0 & \nd 64.5 & \rd 68.4
& \nd 90.7 & \nd 94.3 & \nd 95.4 \\
& \checkmark & - & -
& 12.0
& 93.1 & \nd 96.8 & \rd 97.7
& 93.0 & 96.1 & \rd 96.8
& 48.1 & 59.2 & 62.9
& \rd 50.0 & \rd 63.2 & \nd 69.7
& 89.3 & 93.1 & 94.5 \\
& - & \checkmark & -
& 12.0
& \rd 93.3 & \nd 96.8 & \rd 97.7
& 92.8 & \rd 96.2 & \nd 96.9
& \nd 49.6 & \rd 60.5 & \rd 65.2
& \fs 52.6 & \nd 64.5 & \fs 72.4
& 89.1 & 93.7 & \rd 95.2 \\
& - & - & \checkmark
& 12.6
& \rd 93.3 & \nd 96.8 & \nd 97.8
& \nd 93.2 & \nd 96.4 & \rd 96.8
& 47.4 & 59.9 & 64.6
& \nd 51.3 & \rd 63.2 & 67.1
& \rd 89.8 & \rd 93.8 & 95.1 \\
& \checkmark & \checkmark & \checkmark
& 13.8
& \fs 93.8 & \fs 97.4 & \fs 98.3
& \fs 94.3 & \fs 97.0 & \fs 97.8
& \fs 50.0 & \fs 64.2 & \fs 68.5
& \fs 52.6 & \fs 68.4 & \fs 72.4
& \fs 91.5 & \fs 95.1 & \fs 95.8 \\
\bottomrule
\end{tabular}
\endgroup
\vspace{-0.2cm}
\label{tab:route_ablation_boq}
\end{table*}

\begin{figure}[t]
    \centering
    \includegraphics[width=\columnwidth]{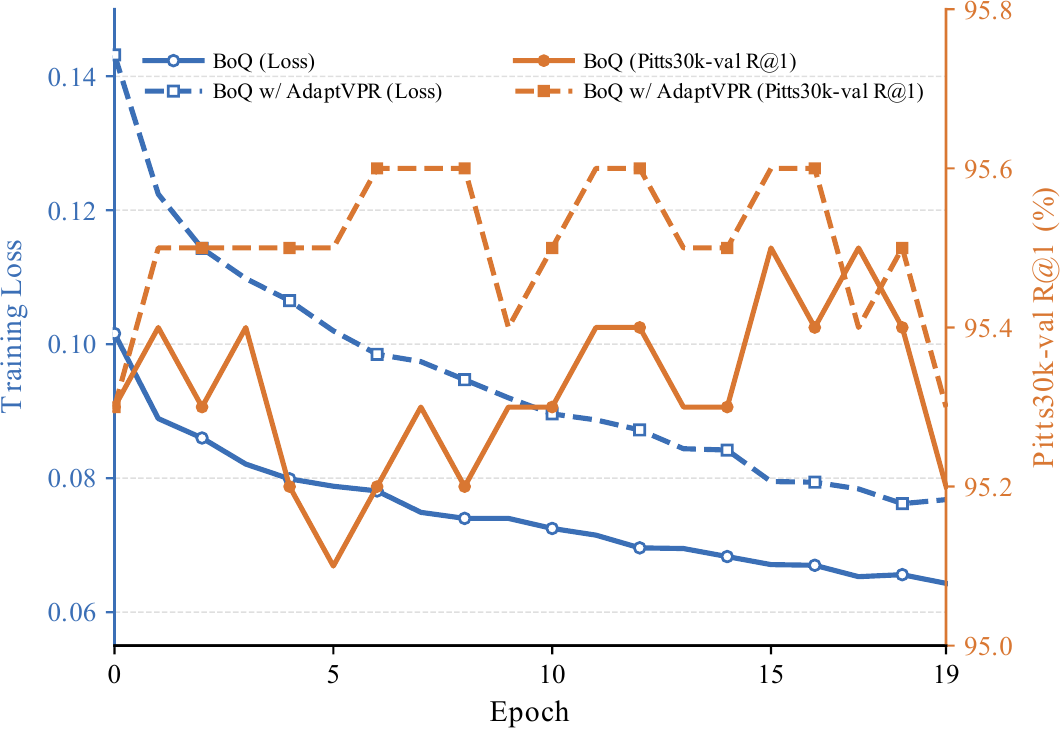}
\caption{
    Training dynamics of BoQ with and without AdaptVPR over the first 20 epochs.
    AdaptVPR produces higher training loss, reflecting increased difficulty from generated same-place hard positives, while optimization remains stable and Pitts30k-val R@1 stays higher during most of training.
    }
    \label{fig:training_dynamics}
    \vspace{-0.2cm}
\end{figure}

Figure~\ref{fig:training_dynamics} compares the training dynamics of BoQ with and without AdaptVPR. After introducing AdaptVPR-generated same-place hard positives, the training loss remains higher, indicating that the model is exposed to more difficult positive pairs rather than simply repeating easy variations of original data. Despite this increased difficulty, the loss decreases steadily throughout optimization, showing that the additional synthetic supervision does not destabilize convergence. Meanwhile, Pitts30k-val R@1 remains higher during most epochs, suggesting that the harder positives provide useful learning signals that translate into improved retrieval performance. These observations complement the benchmark results by showing that AdaptVPR improves training through more challenging positive supervision rather than through changes to the VPR architecture. Since the backbone, aggregation module, descriptor dimensionality, and inference procedure remain unchanged, the additional computational cost is confined to offline generation and verification, without introducing extra retrieval latency or database storage at test time.

\subsection{Ablation Studies}

We conduct a series of ablation studies to analyze the key design choices of AdaptVPR, including quality filtering, route composition, the complete generation procedure, backbone generalization, and synthetic sampling ratio. These experiments examine how different components affect generation effectiveness, training behavior, and retrieval performance.

\textit{1) Effect of Quality Filtering:}
Table~\ref{tab:filtering_ablation_boq_2} evaluates the geometry and diversity filters used to select generated hard positives. Applying either filter alone does not consistently improve retrieval performance, whereas combining both criteria yields the strongest overall results across the evaluated benchmarks. Compared with using all unfiltered synthetic samples, joint filtering improves R@1 by 0.4\% on Pitts30k, 1.2\% on MSLS-val, 0.6\% on SF-XL-Night, 2.6\% on SF-XL-Occlusion, and 0.8\% on SF-XL-v1. Meanwhile, the retained ratio decreases from 100.0\% to 65.6\%, and the average training time is reduced from 14.6 to 13.8 min/epoch. These results indicate that jointly enforcing geometric consistency and appearance diversity is more effective than simply retaining a larger synthetic pool, highlighting the importance of sample quality in generative VPR augmentation.

\textit{2) Effect of Generation Routes:}
Table~\ref{tab:route_ablation_boq} evaluates the contribution of the Global Appearance Route, Local Occlusion Route, and Dual Route. Individual routes provide targeted benefits but are not uniformly optimal across all conditions. For example, the Local Occlusion Route improves SF-XL-Occlusion R@1 from 50.0\% to 52.6\%, confirming its effectiveness for occlusion-heavy scenarios. In contrast, using all three routes yields the strongest overall performance across the evaluated benchmarks. Compared with the unfiltered synthetic setting, the full route combination improves R@1 by 0.4\% on Pitts30k, 1.2\% on MSLS-val, 0.6\% on SF-XL-Night, 2.6\% on SF-XL-Occlusion, and 0.8\% on SF-XL-v1. It also reduces the average training time from 14.6 to 13.8 min/epoch. These results indicate that the three routes provide complementary forms of visual variation, and their combination offers the most effective overall training configuration.

\begin{figure*}[t]
    \centering
    \includegraphics[width=\textwidth]{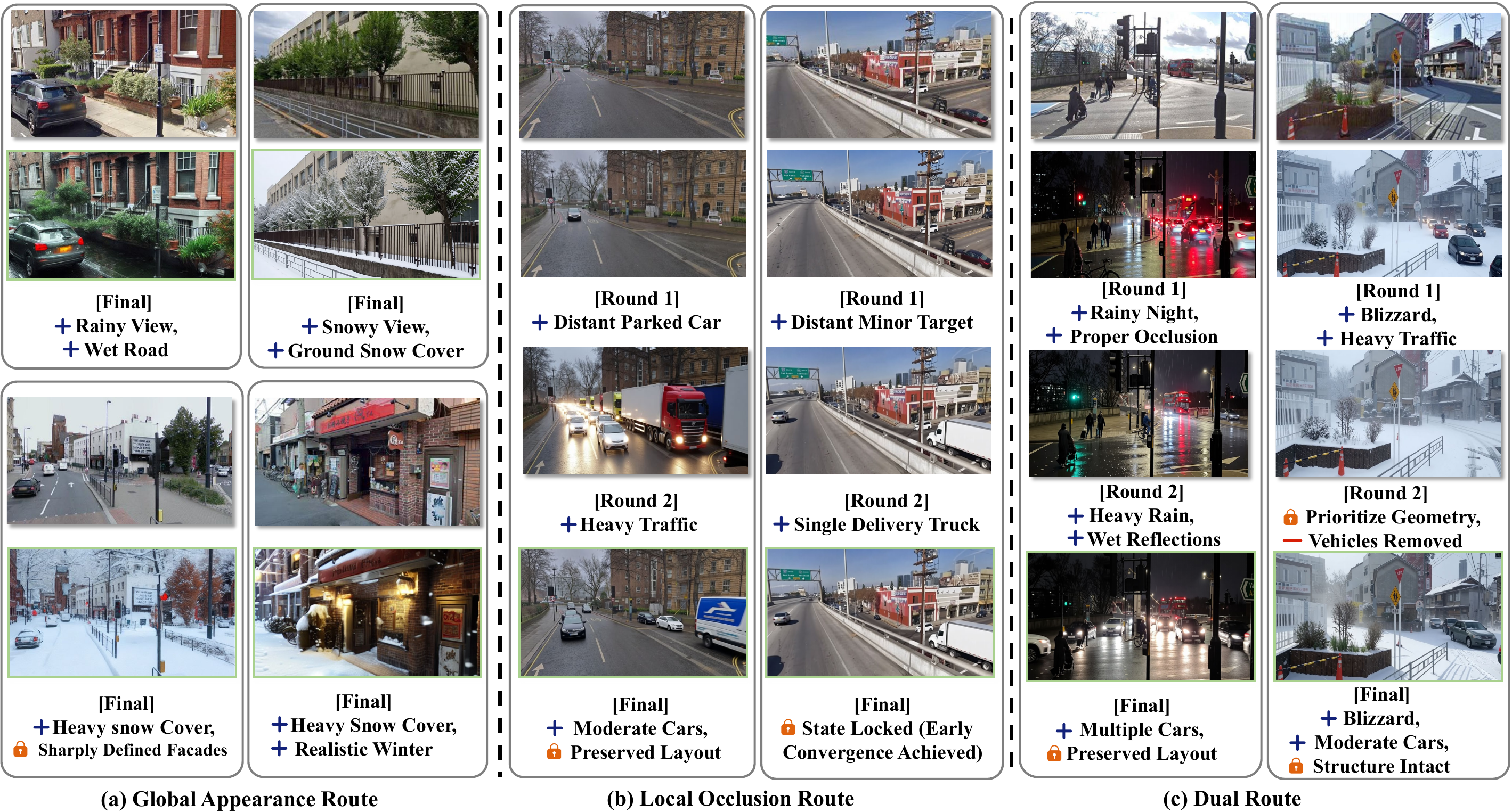}
    \caption{
    Qualitative examples of AdaptVPR under three generation routes.
    (a) The Global Appearance Route introduces scene-level changes in weather, illumination, and time of day through one-shot generation while preserving place structure.
    (b) The Local Occlusion Route introduces dynamic foreground occlusions and refines failed candidates using verification feedback.
    (c) The Dual Route combines global appearance changes with local occlusions and similarly refines failed candidates through verification-guided prompt updates.
    The final accepted samples exhibit challenging visual variations while preserving key structural cues of the original place, providing verified hard positives for VPR training.
    }
    \label{fig:qualitative}
\end{figure*}

To better understand the behavior of AdaptVPR, we present qualitative comparisons in Figure~\ref{fig:qualitative} across three generation strategies: the Global Appearance Route, the Local Occlusion Route, and the Dual Route. The Global Appearance Route focuses on scene level transformations in weather, illumination, and time of day while preserving spatial layout, whereas the Local Occlusion Route introduces object level perturbations (e.g., dynamic vehicles and occlusions) to increase scene complexity without altering the underlying geometry. In contrast, the Dual Route combines both types of modifications in a controlled manner, progressively increasing sample difficulty across multiple rounds under the guidance of a self-reflective agent, while explicitly enforcing structural consistency to produce challenging yet valid synthetic hard positives.

\begin{table}[t]
\centering
\caption{
Analysis of the complete generation procedure in AdaptVPR.
We compare a one-shot baseline with the full AdaptVPR procedure, which combines adaptive route selection, structured prompts, verifier feedback, and repeated generation within the reflection budget.
We report the average number of attempts, end-to-end generation time, and candidate acceptance rate (AR).
}
\label{tab:reflection_extended}
\small
\setlength{\tabcolsep}{1.2pt}
\renewcommand{\arraystretch}{1.1}

\begin{tabular}{@{}lcccccc@{}}
\toprule
\makecell{Generation\\Strategy}
& \makecell{Route\\Selection}
& \makecell{Verifier\\Feedback}
& \makecell{Structured\\Prompt}
& \makecell{Avg.\\Attempts}
& \makecell{Avg.\\Time (s)}
& \makecell{AR\\(\%)} \\
\midrule

One-shot
& Random
& \xmark
& \xmark
& 1.00
& 21.0
& 50.8 \\

AdaptVPR
& Adaptive
& \cmark
& \cmark
& 2.24
& 46.5
& 65.6 \\

\bottomrule
\end{tabular}
\end{table}

\textit{3) Analysis of the Complete Generation Procedure:}
As shown in Table~\ref{tab:reflection_extended}, the complete AdaptVPR procedure increases the candidate acceptance rate from 50.8\% to 65.6\%, indicating that adaptive routing and verifier-guided refinement enable a larger fraction of generated candidates to serve as valid hard positives. This benefit comes with additional generation cost, with the average number of attempts increasing from 1.00 to 2.24 and the average time from 21.0\,s to 46.5\,s. Overall, the results demonstrate a practical trade-off between generation effectiveness and offline computation. Since the two settings differ in routing, prompting, and verifier feedback, this comparison evaluates the complete generation procedure rather than reflection alone.

\begin{table*}[t]
\caption{
Generalization of AdaptVPR across vision foundation backbones.
We compare DINOv2-B and DINOv3-B under their corresponding training and inference resolutions on Pitts30k, MSLS-val, and Tokyo24/7.
}
\small
\centering
\begingroup
\setlength{\tabcolsep}{2.0pt}
\renewcommand{\arraystretch}{0.8}

\begin{tabular}{@{}llccc||ccc||ccc||ccc@{}}
\toprule
\multirow{2}{*}{Method}
& \multirow{2}{*}{Backbone}
& \multirow{2}{*}{Params. (M)}
& \multirow{2}{*}{Train Res.}
& \multirow{2}{*}{Test Res.}
& \multicolumn{3}{c||}{Pitts30k}
& \multicolumn{3}{c||}{MSLS-val}
& \multicolumn{3}{c}{Tokyo24/7} \\
\cline{6-14}
&
&
&
&
&
\rule{0pt}{1.1em}R@1 & R@5 & R@10
& R@1 & R@5 & R@10
& R@1 & R@5 & R@10 \\
\midrule

SALAD
& DINOv2-B
& 88.0
& 224$\times$224
& 322$\times$322
& 92.5 & 96.4 & 97.5
& 92.2 & 96.4 & 97.0
& 94.6 & 97.5 & 97.8 \\

\rowcolor{lightshade}
+AdaptVPR
& DINOv2-B
& 88.0
& 224$\times$224
& 322$\times$322
& 93.2{\scriptsize\textcolor{red}{(+0.7)}} & 96.8 & 97.8
& 93.5{\scriptsize\textcolor{red}{(+1.3)}} & 96.8 & 97.2
& 94.9{\scriptsize\textcolor{red}{(+0.3)}} & 98.1 & 98.4 \\

\midrule

SALAD
& DINOv3-B
& 87.1
& 224$\times$224
& 320$\times$320
& 92.7 & 96.5 & 97.6
& 91.1 & 95.7 & 96.6
& 98.4 & 98.7 & 99.1 \\

\rowcolor{lightshade}
+AdaptVPR
& DINOv3-B
& 87.1
& 224$\times$224
& 320$\times$320
& 93.2{\scriptsize\textcolor{red}{(+0.5)}} & 96.9 & 97.7
& 92.4{\scriptsize\textcolor{red}{(+1.3)}} & 97.2 & 98.0
& 98.7{\scriptsize\textcolor{red}{(+0.3)}} & 98.7 & 99.4 \\

\bottomrule
\end{tabular}
\endgroup
\vspace{-0.2cm}
\label{tab:backbone_generalization}
\end{table*}

\textit{4) Generalization across Vision Foundation Backbones:}
To evaluate the backbone generalization of AdaptVPR, we further conduct experiments on SALAD with DINOv2-B~\cite{dinov2} and DINOv3-B~\cite{dinov3}, as reported in Table~\ref{tab:backbone_generalization}. AdaptVPR consistently improves retrieval performance across both backbones and all three benchmarks. With DINOv2-B, the R@1 gains are 0.7\%, 1.3\%, and 0.3\% on Pitts30k, MSLS-val, and Tokyo24/7, respectively, while the corresponding R@5 gains reach 0.4\%, 0.4\%, and 0.6\%. With DINOv3-B, the R@1 gains are 0.5\%, 1.3\%, and 0.3\%, and MSLS-val further shows gains of 1.5\% and 1.4\% at R@5 and R@10. These consistent improvements across different recall levels indicate that the effectiveness of AdaptVPR transfers well across the evaluated vision foundation backbones.

\begin{table*}[t]
\centering
\caption{
Effect of the real-to-synthetic sampling ratio in AdaptVPR.
The full verified synthetic pool is available in all mixed settings, while only the batch composition is varied.
All settings use the same number of optimization steps.
}
\label{tab:batch_sampling_ratio}

\small
\setlength{\tabcolsep}{4.5pt}
\renewcommand{\arraystretch}{0.95}

\begin{tabular}{@{}c ccc ccc ccc ccc ccc@{}}
\toprule

\multirow{2}{*}{\shortstack{Real:Synthetic\\Batch Ratio}}
& \multicolumn{3}{c}{SF-XL-v1}
& \multicolumn{3}{c}{SF-XL-Night}
& \multicolumn{3}{c}{SF-XL-Occ.}
& \multicolumn{3}{c}{Pitts30k}
& \multicolumn{3}{c}{MSLS-val} \\

\cmidrule(lr){2-4}
\cmidrule(lr){5-7}
\cmidrule(lr){8-10}
\cmidrule(lr){11-13}
\cmidrule(lr){14-16}

& R@1 & R@5 & R@10
& R@1 & R@5 & R@10
& R@1 & R@5 & R@10
& R@1 & R@5 & R@10
& R@1 & R@5 & R@10 \\

\midrule

Real-only
& 89.2 & 93.1 & 94.9
& 47.6 & 59.7 & 64.4
& 51.3 & 63.2 & 68.4
& 93.0 & 96.7 & 97.7
& 93.2 & 96.1 & 96.6 \\

\rowcolor{lightshade}
8:1
& 91.5 & 95.1 & 95.8
& 50.0 & 64.2 & 68.5
& 52.6 & 68.4 & 72.4
& 93.8 & 97.4 & 98.3
& 94.3& 97.0 & 97.8 \\

4:1
& 89.9 & 94.6 & 94.7
& 49.4 & 60.3 & 65.9
& 50.7 & 65.8 & 70.4
& 93.2 & 96.8 & 97.7
& 93.4 & 97.0 & 97.4 \\

2:1
& 88.6 & 94.3 & 95.2
& 48.7 & 59.4 & 65.2
& 50.0 & 61.8 & 67.1
& 93.1 & 96.8 & 97.8
& 92.6 & 95.9 & 96.4 \\

1:1
& 88.4 & 94.3 & 95.3
& 48.9 & 59.7 & 64.6
& 51.3 & 61.8 & 67.1
& 93.1 & 96.7 & 97.7
& 92.7 & 95.9 & 96.5 \\

\bottomrule
\end{tabular}
\vspace{-0.3cm}
\end{table*}

\textit{5) Effect of Synthetic Sampling Ratio:}
To examine the effect of the synthetic-data ratio on VPR training, we vary the real-to-synthetic ratio within each mini-batch while keeping the verified synthetic pool and the number of optimization steps fixed. As shown in Table~\ref{tab:batch_sampling_ratio}, the 8:1 setting provides the best overall balance and achieves the highest R@1 across all evaluated benchmarks. Compared with real-only training, R@1 increases from 93.0\% to 93.8\% on Pitts30k and from 93.2\% to 94.3\% on MSLS-val. 
On the SF-XL benchmarks, R@1 improves by 2.3\%, 2.4\%, and 1.3\% on SF-XL-v1, SF-XL-Night, and SF-XL-Occlusion, respectively.
Increasing the synthetic sampling frequency beyond the 8:1 setting does not yield consistent additional gains; for example, R@1 on SF-XL-v1 decreases to 89.9\% at 4:1 and falls slightly below the real-only baseline at 2:1 and 1:1. These results indicate that moderate exposure to verified hard positives provides a better balance between enriching appearance variation and preserving the real-data distribution.

Figs.~\ref{fig:additional_verification_global_local} and~\ref{fig:additional_verification_dual} present verification examples across the three routes. The first shows Global Appearance and Local Occlusion examples, while the second focuses on the Dual Route. Each example visualizes the variation with geometric correspondences and verification scores $s_{\rm geo}$ and $s_{\rm div}$.

\section{Discussions}
\label{sec:discussion}

The experimental results indicate that AdaptVPR improves robustness by enriching the appearance variations observed for the same place during training. Across different VPR baselines and vision foundation backbones, AdaptVPR consistently improves retrieval performance on standard benchmarks, with substantially larger gains under challenging domain shifts. This trend suggests that generated hard positives are particularly valuable when test conditions are insufficiently represented in the original training data. The ablation results further show that effective generative augmentation depends not only on the amount of synthetic data, but also on how the samples are generated, verified, and incorporated into training.

\noindent\textbf{Data Quality and Sampling.}
The complementary generation routes cover different forms of visual change, while geometry and diversity verification reduce the risk of introducing invalid positives. Meanwhile, the batch sampling study shows that increasing synthetic exposure does not monotonically improve performance. A moderate real to synthetic ratio achieves a better overall tradeoff, suggesting that the value of synthetic data lies in informative and well controlled hard positives rather than simply increasing their quantity.

\noindent\textbf{Practicality and Limitations.}
AdaptVPR operates entirely at the training data level and therefore introduces no additional inference module, descriptor dimension, or retrieval storage cost. Its main overhead comes from offline generation, verification, and prompt refinement. In addition, the geometric consistency score remains a proxy for place identity and cannot fully guarantee that every generated image preserves the original location. The current framework also relies on predefined generation routes and verification thresholds, which may limit its flexibility under unseen environments.

\noindent\textbf{Future Directions.}
Future work could explore more adaptive route scheduling and verification strategies, including stronger place-consistency signals and more efficient feedback mechanisms. Another promising direction is to reduce the computational cost of synthetic data construction through lightweight generation, early rejection, or selective refinement. Beyond urban street-view imagery, AdaptVPR could also be extended to broader geographic environments and other sensing modalities~\cite{LaVPR,han2025multimodal,Mms-vpr}.

\section{Conclusion}
\label{sec:conclusion}

In this work, we proposed AdaptVPR, a route-aware generative augmentation framework for constructing same-place hard positives to improve VPR robustness under domain shift. AdaptVPR combines vision-language scene understanding, rule-based route scheduling, complementary Global Appearance, Local Occlusion, and Dual Routes, together with geometry and diversity verification and selective reflection. By generating challenging appearance variations while controlling structural drift, AdaptVPR expands the diversity of same-place observations available during training. Using this framework, we construct AdaptCities with 160K verified synthetic same-place hard positives.
Extensive experiments across diverse VPR baselines and vision foundation backbones demonstrate consistent improvements on standard benchmarks and larger robustness gains under challenging seasonal, nighttime, and occlusion conditions, with R@1 improvements of up to 9.2\%. The consistent gains across heterogeneous VPR models further indicate that the effectiveness of AdaptVPR is not tied to a particular backbone or feature aggregation design.
Overall, AdaptVPR provides a general training data augmentation strategy for improving VPR robustness under domain shift.

\clearpage

\begin{strip}
\centering
\includegraphics[width=\textwidth, keepaspectratio]{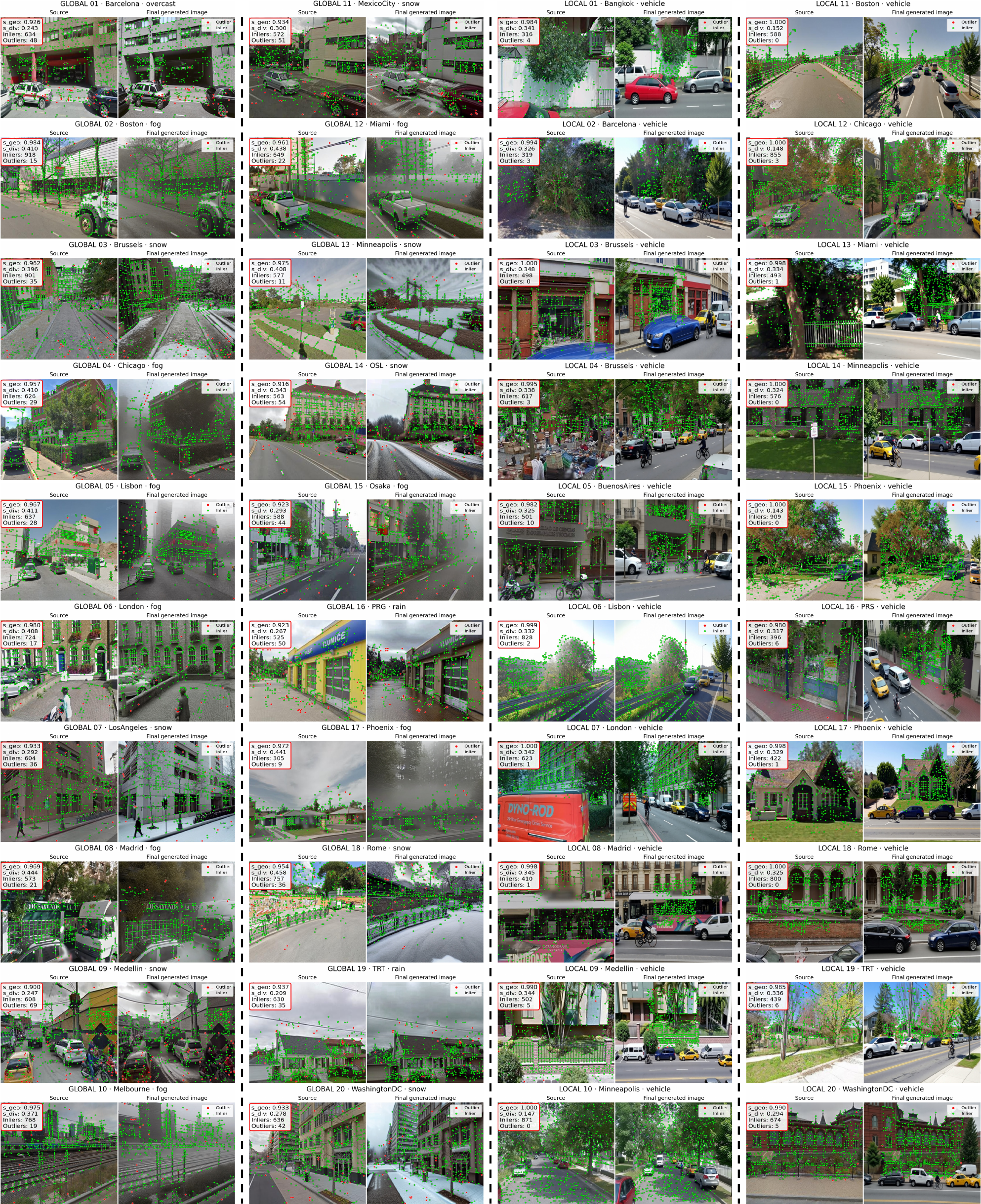}

\captionof{figure}{
Additional qualitative verification examples from the Global Appearance and Local Occlusion Routes. Each example shows the source and generated images together with local feature correspondences, geometric consistency score $s_{\rm geo}$, appearance diversity score $s_{\rm div}$, and geometric inlier and outlier counts. The examples demonstrate substantial appearance or visibility changes while preserving structural correspondence with the reference scene.
}
\label{fig:additional_verification_global_local}
\end{strip}

\clearpage

\begin{strip}
\centering
\includegraphics[width=\textwidth, keepaspectratio]{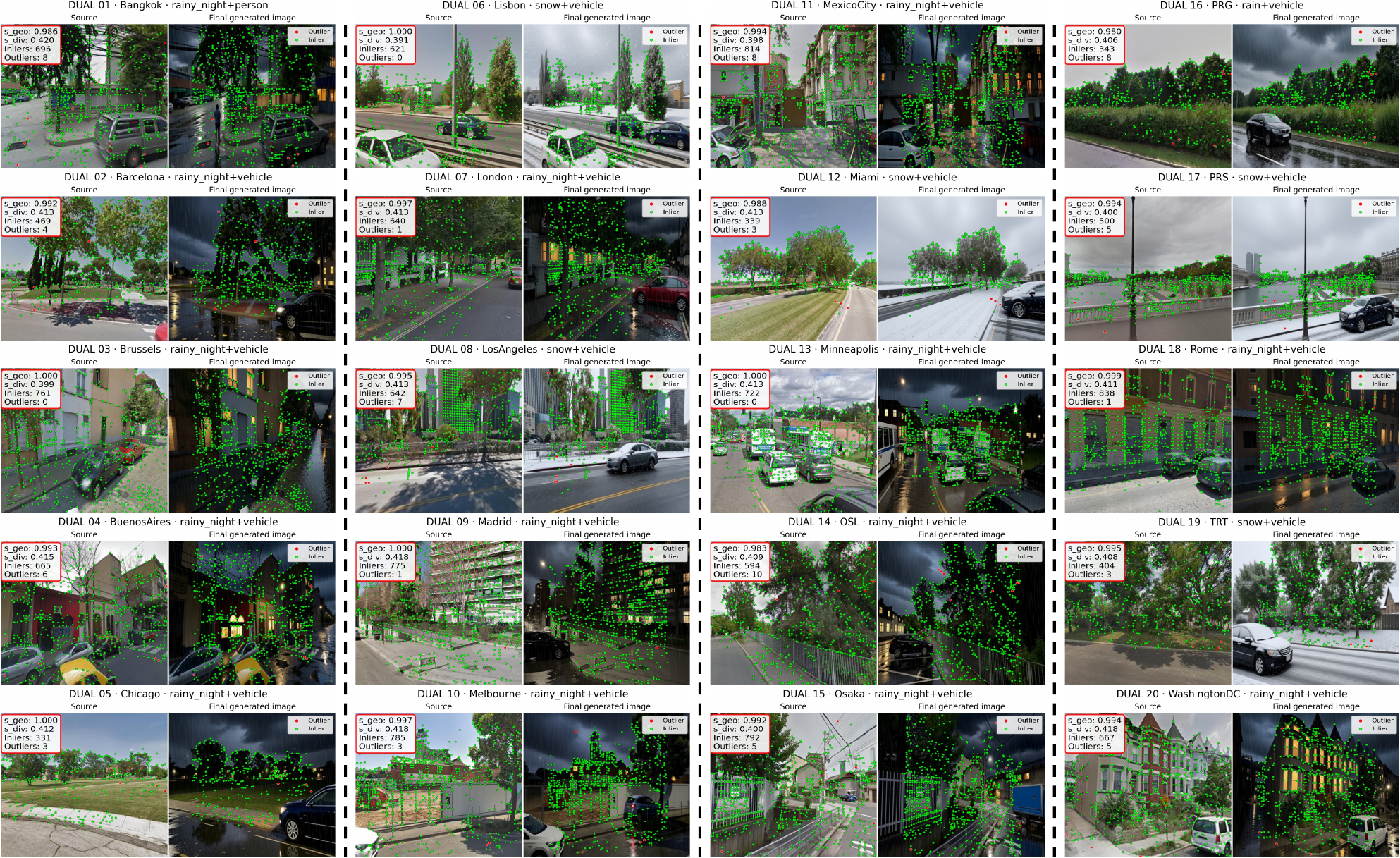}

\captionof{figure}{
Additional qualitative verification examples from the Dual Route. These examples combine local occlusions with global appearance changes such as rain, snow, illumination variation, and nighttime conditions. The reported $s_{\rm geo}$ and $s_{\rm div}$ scores, together with geometric inlier and outlier counts, demonstrate challenging compound variations while maintaining structural correspondence with the reference scene.
}
\label{fig:additional_verification_dual}
\end{strip}

\section*{Acknowledgments}
This work was supported by the Beijing Natural Science Foundation (No. JQ23014) and the National Natural Science Foundation of China (No. 62271074); and in part by the BUPT Kunpeng\&Ascend Center of Cultivation, the Taishan Scholars Program (No. TSQN202507241), the Key R\&D Program of Shandong Province, China (No. 2025KJHZ013), the Shandong Provincial University Youth Innovation and Technology Support Program (No. 2022KJ291), the Shandong Provincial Natural Science Foundation for Young Scholars Program (No. ZR2025QC1627), and the Qilu University of Technology (Shandong Academy of Sciences) Youth Outstanding Talent Program (No. 2024QZJH02).

\ifCLASSOPTIONcaptionsoff
  \newpage
\fi

\bibliographystyle{IEEEtran}
\bibliography{IEEEabrv,reference}

\end{document}